\documentclass[11pt,reqno]{amsart}
\usepackage{graphicx,enumerate,amssymb,bbold}
\usepackage[notrig]{physics}
\usepackage{tabularx}
\usepackage{multirow}
\usepackage[font=footnotesize,labelfont=small]{subcaption}
\usepackage[left=2.5cm,right=2.5cm,top=2.5cm,bottom=2.5cm,includeheadfoot]{geometry} 
\usepackage[multidot]{grffile}
\usepackage[colorlinks=true, pdfstartview=FitV, linkcolor=blue, 
            citecolor=blue, urlcolor=blue]{hyperref}
            
\usepackage{mathtools}

\usepackage{tikz}
\usetikzlibrary{graphs,positioning,fit,calc,arrows.meta}

\numberwithin{equation}{section}
\numberwithin{figure}{section}
\theoremstyle{plain}

\theoremstyle{definition}

\newtheorem*{remark*}{Remark}

\newcommand{\bitem}{\begin{itemize}}
\newcommand{\eitem}{\end{itemize}}
\newcommand{\mc}[1]{\mathcal{#1}}

\newcommand{\N}{\mathbb{N}}
\newcommand{\R}{\mathbb{R}}

\newcommand{\EE}{\mathbb{E}}

\newcommand{\Z}{\mathbb{Z}}

\newcommand{\bpm}{\begin{pmatrix}}
\newcommand{\epm}{\end{pmatrix}}
\newcommand{\bvm}{\begin{vmatrix}}
\newcommand{\evm}{\end{vmatrix}}
\newcommand{\bsm}{\left(\begin{smallmatrix}}
\newcommand{\esm}{\end{smallmatrix}\right)}
\newcommand{\T}{\top}

\newcommand{\ol}[1]{\overline{#1}}
\newcommand{\la}{\langle}
\newcommand{\ra}{\rangle}

\newcommand{\mrm}[1]{\mathrm{#1}}

\newcommand{\mfk}[1]{\mathfrak{#1}}

\newcommand{\veps}{\varepsilon}

\newcommand{\gdw}{\Leftrightarrow}

\newcommand{\eins}{\mathbb{1}}

\newcommand{\LG}[1]{\mathrm{#1}}

\DeclareMathSymbol{\mydiv}{\mathbin}{symbols}{"04}

\DeclareMathOperator{\Diag}{Diag}

\DeclareMathOperator{\dom}{dom}
\DeclareMathOperator{\intr}{int}
\DeclareMathOperator{\rint}{rint}

\DeclareMathOperator*{\argmin}{arg\, min}
\DeclareMathOperator*{\argmax}{arg\, max}

\DeclareMathOperator{\vecmin}{vecmin}
\DeclareMathOperator{\sinc}{sinc}

\DeclareMathOperator{\ggrad}{grad}

\DeclareMathOperator{\Log}{Log}
\DeclareMathOperator{\Exp}{Exp}

\DeclareMathOperator{\expm}{expm}
\DeclareMathOperator{\logm}{logm}
\newcommand{\bigslant}[2]{{\protect\raisebox{.2em}{$#1$}\big/\protect\raisebox{-.2em}{$#2$}}}
\newcommand{\coloneqq}{\stackrel{\text{\tiny def}}{=}}

\title[Unsupervised Assignment Flow]{Unsupervised Assignment Flow: Label Learning on Feature Manifolds \\ by Spatially Regularized Geometric Assignment}
\author[A.~Zern, M.~Zisler, S.~Petra, C.~Schn\"{o}rr]{Artjom Zern, Matthias Zisler, Stefania Petra, Christoph Schn\"{o}rr}
\address[A.~Zern]{Image and Pattern Analysis Group, Heidelberg University, Germany} 
\email{artjom.zern@iwr.uni-heidelberg.de}
\address[M.~Zisler]{Image and Pattern Analysis Group, Heidelberg University, Germany} 
\email{zisler@math.uni-heidelberg.de}
\address[S.~Petra]{Mathematical Imaging Group, Heidelberg University, Germany} 
\email{petra@math.uni-heidelberg.de}
\urladdr{\url{https://www.stpetra.com}}
\address[C.~Schn\"{o}rr]{Image and Pattern Analysis Group, Heidelberg University, Germany} 
\email{schnoerr@math.uni-heidelberg.de}
\urladdr{\url{https://ipa.math.uni-heidelberg.de}}
\date{} 

\thanks{This work is supported by Deutsche Forschungsgemeinschaft (DFG) under Germanys Excellence Strategy EXC-2181/1 - 390900948 (the Heidelberg STRUCTURES Excellence Cluster) and by the Research Training Group funded by the DFG, Grant GRK 1653. \\
This is a pre-print of an article published in Journal of Mathematical Imaging and Vision. The final authenticated version is available online at: \url{https://doi.org/10.1007/s10851-019-00935-7}}


\begin{document}

\begin{abstract}
This paper introduces the \textit{unsupervised assignment flow} that couples the assignment flow for supervised image labeling \cite{Astrom:2017ac} with Riemannian gradient flows for label evolution on feature manifolds. The latter component of the approach encompasses extensions of state-of-the-art clustering approaches to manifold-valued data. Coupling label evolution with the spatially regularized assignment flow induces a sparsifying effect that enables to learn compact label dictionaries in an unsupervised manner. Our approach alleviates the requirement for supervised labeling to have proper labels at hand, because an initial set of labels can evolve and adapt to better values while being assigned to given data. The separation between feature and assignment manifolds enables the flexible application which is demonstrated for three scenarios with manifold-valued features. Experiments demonstrate a beneficial effect in both directions: adaptivity of labels improves image labeling, and steering label evolution by spatially regularized assignments leads to proper labels, because the assignment flow for supervised labeling is exactly used without any approximation for label learning.
\end{abstract}

\maketitle
\tableofcontents

\section{Introduction}
\label{sec:Introduction}

\subsection{Motivation.} Geometric methods based on manifold models of data and Riemannian geometry are nowadays widely employed in image processing and computer vision \cite{Turaga:2016aa}. For example, covariance descriptors play a prominent role \cite{Turaga:2016aa,Cherian:2016aa}. Covariance descriptors are typically applied to the detection and classification of entire images (e.g.~faces, texture) or videos (e.g.~action recognition), or as descriptors of local image structure. An important task in this context is to compute a \textit{codebook} of covariance descriptors that can be used for solving a task at hand like, e.g., image classification by nearest-neighbor search \cite{Cherian:2013aa}, or  image labeling \cite{Kappes:2015aa} using the codebook descriptors as labels. 

\begin{figure}[ht]
\begin{center}
\begin{tabular}{cc@{\hskip0.4em}ccc}
\footnotesize \textbf{input data} & \shortstack{\footnotesize \textbf{clustering \emph{only}} \\ \footnotesize \textbf{in feature space}} & \shortstack{\footnotesize \textbf{spatially regularized} \\ \footnotesize \textbf{label assignment}} & \footnotesize \textbf{\textit{proposed approach}} & \footnotesize \textbf{ground truth} \\
\includegraphics[width=0.17\textwidth]{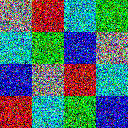} &
\includegraphics[width=0.17\textwidth]{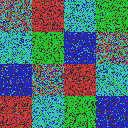} &
\includegraphics[width=0.17\textwidth]{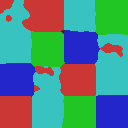} &
\includegraphics[width=0.17\textwidth]{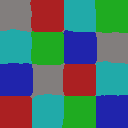} &
\includegraphics[width=0.17\textwidth]{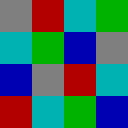} \\
& \multicolumn{2}{c}{\includegraphics[width=0.15\textwidth]{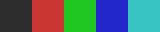}} &
\includegraphics[width=0.15\textwidth]{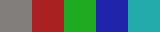} &
\includegraphics[width=0.15\textwidth]{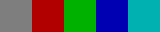}
\end{tabular}
\parbox{0.18\textwidth}{\centering\small (a)}
\parbox{0.18\textwidth}{\centering\small (b)}
\parbox{0.18\textwidth}{\centering\small (c)}
\parbox{0.18\textwidth}{\centering\small (d)}
\parbox{0.18\textwidth}{\centering\small (e)}
\end{center}
\caption{\textbf{Importance of spatially regularized assignments for label \textit{learning}.} (a) Input data: a synthetic image corrupted by Gaussian noise. (b) + (c) The classical two-step approach of  clustering in feature space first (panel b) followed by \textit{supervised} label assignment (panel c) performs poorly, despite spatial regularization. (d) By \textit{coupling} label evolution and spatially regularized assignment both the label set and the labeled image can be drastically improved. (e) Ground truth labeling and label set. Label sets resulting from (b), (d) and (e) are depicted below the respective image labeling results.} \label{fig:motivation}
\end{figure}

The recent work \cite{Harandi2016} introduced a state-of-the-art method for computing such codebooks. After embedding descriptors into a reproducing kernel Hilbert space  \cite{Hofmann:2008aa}, given data are  approximated by a kernel expansion, and a sparse subset can be determined by $\ell_{1}$-regularization of the expansion coefficients. This method works \textit{entirely in feature space}, however, and ignores the \textit{spatial structure of codebook assignments} to data, which is unfavorable in connection with image labeling. Figure~\ref{fig:motivation} illustrates why the spatial structure of label assignments should also drive the \textit{evolution of labels} in feature space for \textit{unsupervised} label learning, if the resulting labels are subsequently used for \textit{supervised} image labeling for which spatial regularization is typically enforced as well. 

We show in this paper how the approach of \cite{Astrom:2017ac} to spatially regularized label assignment can be combined with basic clustering approaches after extending the latter to feature manifolds, to perform \textit{unsupervised label learning} from manifold-valued feature data through spatially regularized label assignment. Our approach is \textit{consistent and natural} in that the  \textit{very same} approach \cite{Astrom:2017ac} for supervised image labeling is also used for the unsupervised learning of proper labels for this task.

\subsection{Related Work, Contribution} The classical approach for the unsupervised learning of feature prototypes (`labels') is the mean-shift iteration \cite{Fukunaga:1975aa,Comaniciu:2002aa}, which iteratively seeks modes (local peaks) of the feature density distribution through the averaging of features within local neighborhoods. This has been generalized to \textit{manifold-valued} features by 
\cite{Subbarao:2009aa}, by replacing ordinary mean-shifts by Riemannian (Fr\'echet, Karcher) means \cite{Karcher:1977aa}. The common way to take into account the \textit{spatial structure} of label assignments is to \textit{augment} the feature space by \textit{spatial coordinates}, e.g.~by turning a color feature $(r,g,b)$ into the feature vector $(x,y,r,g,b)$. This \textit{merge} of feature space and spatial domain has a conceptual drawback, however: the \textit{same} color vector $(\ol{r},\ol{g},\ol{b})$ observed at two \textit{different} locations $(x_{1},y_{1},\ol{r},\ol{g},\ol{b})$, $(x_{2},y_{2},\ol{r},\ol{g},\ol{b})$ defines two \textit{different} feature vectors, and hence, these two feature vectors may be assigned to different prototypes during clustering despite containing the same color information. Furthermore, clustering spatial coordinates into \textit{centroids} by mean-shifts (together with the features) \textit{differs} from \textit{unbiased spatial} regularization as performed by variational approaches, graphical models or the assignment flow approach of \cite{Astrom:2017ac}, where regularization does \textit{not} depend on the location of centroids and the corresponding shape of local density modes.

\vspace{0.1cm}
\noindent
We introduce a novel approach which has the following properties:
\begin{enumerate}[(i)]
\item 
The approach incorporates and performs \textit{unsupervised learning of manifold-valued features}, henceforth called \textit{labels}. The approach applies to any feature manifold \cite{Subbarao:2009aa} for which the corresponding operations defined below like, e.g., Riemannian means are well-defined and computationally feasible. Experiments using $S^{1}$-valued data (2D orientations), $\LG{SO}(3)$-valued data (orthogonal frames) and features on the positive definite matrix manifold (covariance descriptors) illustrate our approach.
\item
The evolution of labels (unsupervised learning) is driven by \textit{spatially regularized assignments} which are \textit{not} biased toward any spatial centroids. This is accomplished by applying the smooth geometric assignment approach to image labeling recently introduced by \cite{Astrom:2017ac}.
\item
The smooth settings of both (i), (ii) enable to define a \textit{smooth coupled flow}
\begin{equation}\label{eq:def-GW-flow}
(\dot M, \dot W) = \mc{V}(M,W)
\end{equation}
where $\dot M$ denotes the evolution of labels and $\dot W$ the evolution of spatially regularized label assignments that \textit{interact} through a coupling vector field $\mc{V}$.
Concrete instances of \eqref{eq:def-GW-flow} are \eqref{eq:coupled-a}, \eqref{eq:coupled-b} and \eqref{eq:UAF}.
This interaction keeps both domains (i) and (ii) \textit{separate} and hence enables to apply flexibly our approach to various feature manifolds, using the \textit{same} regularized assignment mechanism. 
\end{enumerate}
A preliminary version of this paper \cite{Zern:2018ac} introduced the approach called `coupled flow A' in this paper. The present paper elaborates this conference paper in many ways as illustrated by Figure~\ref{fig:structure}, including a generalization to a one-parameter family of coupled flows and a more comprehensive experimental evaluation. 
\subsection{Organization}
\begin{figure}[ht]
\begin{center}
\resizebox{\textwidth}{!}{
\begin{tikzpicture}[
	node distance=4em,
	mynode/.style={rectangle, draw=orange, fill=yellow, fill opacity=0.2, draw opacity=1, text opacity=1},
	every edge/.append style={->, >=latex, line width=1.5pt},
]
	\node[mynode] (C1) {soft $k$-means (Sec.~\ref{sec:Euclidean-k-Means})};
	\node[mynode] (C2) [right=4.5em of C1]{divergence-based EM (Sec.~\ref{eq:Divergences-EM})};
	\node[mynode] (C3) [right=4.5em of C2] {$k$-center clustering (Sec.~\ref{sec:Metric-Clustering})};
	\node[mynode] (CM1) [below=of C1] {soft $k$-means on $\mathcal{M}$ (Sec.~\ref{sec:soft-k-manifold})};
	\node[mynode] (CM2) [below=of C2] {divergence-based EM on $\mathcal{M}$ (Sec.~\ref{sec:manifold-EM})};
	\node[mynode] (assignment) [below=of $(CM1)!0.5!(CM2)$] {assignment flow (Sec.~\ref{sec:Assignment-Supervised})};
	\node (UAFText) [below=5.75em of assignment, inner sep=0em] {unsupervised assignment flow (Sec.~\ref{sec:coupled-flow-final})};
	\node[mynode] (CFa) [below=8em of CM1] {coupled flow A (Sec.~\ref{sec:regularized-soft-k-means-manifold})};
	\node[mynode] (CFb) [below=8em of CM2] {coupled flow B (Sec.~\ref{sec:regularized-EM-manifold})};
	\node[mynode, fit=(UAFText)(CFa)(CFb), inner sep=0.75em] (UAF) {};
	
	\node[rectangle, draw=black, anchor=south east, thick] at (C3.east |- UAF.south) { 
		\begin{tabular}{l@{\hskip6pt}c@{\hskip6pt}l}
		\tikz[baseline=-0.5ex]{\draw[red!50!black] (0,0) edge[->, >=latex, line width=1.5pt] (1,0);} & : & adaption of method \\
		\tikz[baseline=-0.5ex]{\draw[green!50!black, densely dashed] (0,0) edge[->, >=latex, line width=1.5pt] (1,0);} & : & preprocessing for
		\end{tabular}
	};

	\draw[red!50!black] (C1) edge (CM1);
	\draw[red!50!black] (C2) edge (CM2);
	\draw[red!50!black] (CM1) edge (CFa);
	\draw[red!50!black] (assignment) edge (CFa);
	\draw[red!50!black] (CM2) edge (CFb);
	\draw[red!50!black] (assignment) edge (CFb);
	\draw[green!50!black, densely dashed] (C3) edge[in=30,out=-90] (UAF.east);
\end{tikzpicture}
}
\end{center}
\caption{
\textbf{Organization of this paper:}
Three basic clustering algorithms are summarized in Section~\ref{sec:Preliminaries-Clustering}. To of them, \textit{soft-$k$-means clustering} and \textit{divergence-based EM-iteration}, are generalized to feature data taking values in a Riemannian manifold $\mc{M}$ and coupled with the (spatially regularized) \textit{assignment flow}. A smooth interpolation of the resulting \textit{coupled flow A} and \textit{coupled flow B} finally defines the \textit{unsupervised assignment flow}.
}
\label{fig:structure}
\end{figure}
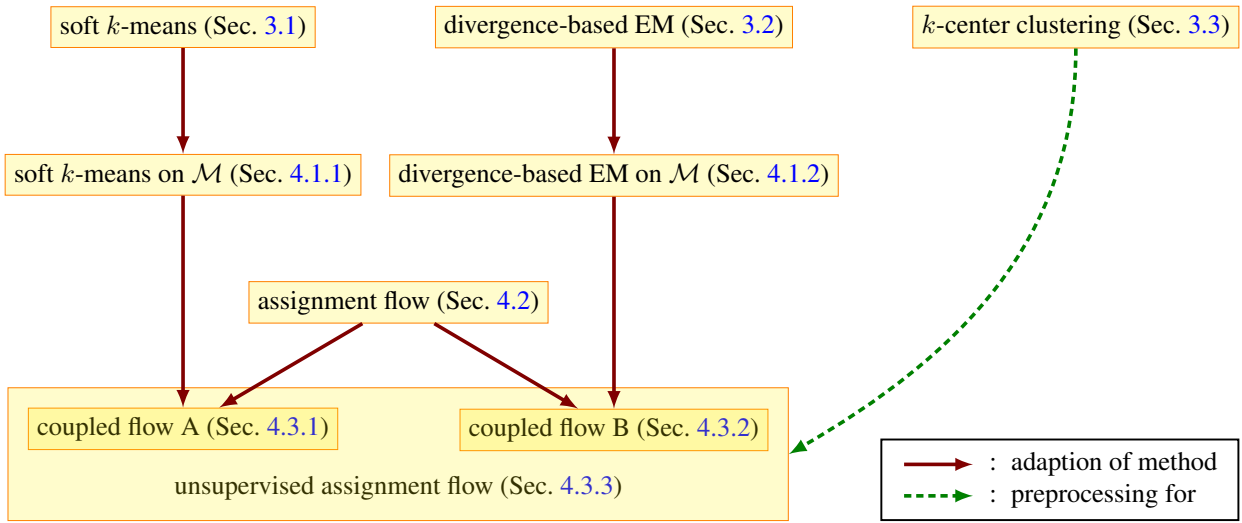

After introducing notation from differential geometry and providing mathematical background on divergence functions in Section~\ref{sec:Preliminaries}, three basic clustering concepts are summarized in Section~\ref{sec:Preliminaries-Clustering}. 
While greedy-based $k$-center clustering (Section~\ref{sec:Metric-Clustering}) will only serve  as a preprocessing step, soft-$k$-means clustering (Section~\ref{sec:Euclidean-k-Means}) and divergence-based EM-iteration (Section~\ref{eq:Divergences-EM}), which perform classical label evolution by mean-shift iteration, will be building blocks for subsequent methods (see Figure~\ref{fig:structure}). 
First, these two methods are adjusted to manifold-valued data (Section~\ref{sec:soft-k-manifold} and \ref{sec:manifold-EM}), and then each of them is modified and coupled with the assignment flow (Section~\ref{sec:Assignment-Supervised}) that induces a sparsifying effect through spatial regularization.
Coupling with soft-$k$-means leads to the `coupled flow A' (Section~\ref{sec:regularized-soft-k-means-manifold}), while coupling with EM-iteration leads to the new `coupled flow B' (Section~\ref{sec:regularized-EM-manifold}). Afterward, we provide a more general natural definition of a one-parameter family of \textit{unsupervised assignment flows} (Section~\ref{sec:coupled-flow-final}) that smoothly interpolates both coupled flows and includes them as special cases. 
Numerical integration of the unsupervised assignment flow is discussed in Section~\ref{sec:geoNumInt}.
Section~\ref{sec:Feature-Manifolds} deals with particular feature manifolds that will be used as case studies and numerical experiments in Section~\ref{sec:Experiments}.

\subsection{Basic Notation} We set $[n]=\{1,2,\dotsc,n\}$ for $n \in \N$ and $\eins = (1,1,\dotsc,1)^{\T}$ with dimension depending on the context. Euclidean vectors are enumerated by superscripts with components indexed by subscripts $x^{i} = (x^{i}_{1},\dotsc,x^{i}_{d})^{\T} \in \R^{d}$. $\la \cdot,\cdot \ra$ denotes the Euclidean inner product $\la u,v \ra = \sum_{i \in [d]} u_{i} v_{i}$ of two vectors or the Frobenius inner product $\la A, B \ra = \tr(A^{\T} B)$ for matrices. Throughout the paper, the symbols $I, J$ denote 
\begin{equation}
\begin{aligned}
I &\colon\text{set of data indices}, \\
J &\colon\text{set of label indices},
\end{aligned}
\end{equation}
with cardinalities $|I|$ and $|J|$. The relation $A \succ 0$ ($A \succeq 0$) indicates that a symmetric matrix $A=A^{\T}$ is positive (semi-) definite. The $(d-1)$-dimensional probability simplex is denoted by
\begin{equation}
\Delta_{d} = \big\{x \in \R^{d} \colon x_{i}\geq 0,\, i \in [d],\, \la\eins,x\ra=1\big\} \subset \R^{d}.
\end{equation}
For \textit{strictly} positive probability vectors $0 < p, q \in \Delta_{n}$, we denote \textit{componentwise} multiplication and division efficiently by $p q$ and $\frac{p}{q}$, respectively. 
It will be convenient to denote the exponential function with vectors as argument in two alternative ways,
\begin{equation}\label{eq:def-exp-function}
\exp(x) = e^{x} \coloneqq (e^{x_{1}},\dotsc,e^{x_{d}})^{\T},
\end{equation}
and similarly with $\log(x)$ and strictly positive vectors $x > 0$. $(\mc{M},g)$ \textit{generally} denote some Riemannian manifold $\mc{M}$ with metric $g$, whereas $\mc{S}, \mc{W}, \mc{P}_{s}$ denote \textit{specific} Riemannian manifolds defined in the subsequent sections. In this context,
the symbols
\begin{alignat*}{3}
&\exp_{\mc{M},p} &\colon& \text{exponential map of manifold $\mc{M}$ -- cf. \eqref{eq:exp-Riemannian}} \\
&\Exp_{p} &\colon& \text{$\alpha$-expontial map (with $\alpha=1$) of the open simplex $\mc{S}$ -- cf. \eqref{eq:def-exp-S}} \\
&\exp_{p} &\colon& \text{lifting map onto $\mc{S}$ -- cf. \eqref{eq:def-exp-1}}
\end{alignat*}
\textit{with} subscript denote (exponential) maps of $\mc{M}$ or $\mc{S}$ in particular and should not be confused with the exponential function \eqref{eq:def-exp-function} that is uniquely denoted \textit{without} subscript.
\section{Background}
\label{sec:Preliminaries}

This section collects background material required in this paper. Section \ref{sec:Preliminaries-Geometry} covers basic notion of differential geometry. We recommend \cite{Lee:2013aa,Jost:2017aa} for further reading. Section \ref{sec:Divergence-Functions} recalls the notion of a \textit{divergence function}. Such functions are used in applications in lieu of the squared Riemannian distance if evaluating the latter is computationally too involved. See \cite{Basseville:2013aa} for a survey and \cite{Amari:2010aa} for a mathematical account. Concrete divergence functions are considered in Section~\ref{sec:Feature-Manifolds}.

\subsection{Basic Notions from Differential Geometry}
\label{sec:Preliminaries-Geometry}
Let $(\mc{M},g)$ be a Riemannian manifold with metric $g$. 
We denote the tangent and cotangent space at $p \in \mc{M}$ by $T_p \mc{M}$ and $T_p^* \mc{M}$. $\mc{F}(\mc{M})$ denotes the set of smooth functions $f \colon \mc{M} \to \R$, and 
$\mfk{X}(\mc{M})$ denotes the set of all smooth vector fields, i.\,e. smooth sections $X \colon \mc{M} \rightarrow T\mc{M}$ of the tangent bundle $T\mc{M}$. Subscripts $X_{p} \in T_{p}\mc{M}$ indicate the evaluation of a vector field $X \in \mfk{X}(\mc{M})$. Further, $\mfk{X}^{\ast}(\mc{M})$ denotes the set of all smooth covector fields (i.\,e. one-forms).
$df \in \mfk{X}^{\ast}(\mc{M})$ denotes the differential of a function $f \in \mc{F}(\mc{M})$ and $df(X)$ and $df_{p}(v)$ its action on $X \in \mfk{X}(\mc{M})$ and $v \in T_{p}\mc{M}$. We use both notations
\begin{equation}
g(X,Y) = \la X, Y \ra_{g},\quad X, Y \in \mfk{X}(\mc{M})
\end{equation}
when evaluating the metric. 
The \textit{Riemannian gradient} of a function $f \in \mc{F}(\mc{M})$ is the vector field 
\begin{subequations}\label{eq:def-grad}
\begin{align}\label{eq:def-grad-1}
\ggrad f &\in \mfk{X}(\mc{M})
\intertext{
defined by
}\label{eq:def-grad-2}
\la\ggrad f, X\ra_{g} 
&= df(X) = X f,\qquad\forall X \in \mfk{X}(\mc{M}).
\end{align}
\end{subequations}
Let $\widehat g$ denote the linear tangent-cotangent isomorphism\footnote{The maps $\widehat{g}$ and $\widehat{g}^{-1}$ are sometimes denoted with $\flat$ and $\sharp$ in the literature (`musical isomorphism'). We stick to the notation from \cite{Lee:2013aa} here.}
\begin{equation}\label{eq:def-grad-3}
\widehat g \colon \mfk{X}(\mc{M}) \to \mfk{X}^{\ast}(\mc{M}),\qquad
\widehat g(X)(Y) \coloneqq g(X,Y),\quad
\forall X, Y \in \mfk{X}(\mc{M})
\end{equation}
that associates with a vector field $X$ the covector field $\widehat g(X) = g(X,\cdot)$. Then by \eqref{eq:def-grad-2},
\begin{equation}\label{eq:def-grad-4}
\ggrad f = \widehat g^{-1}(df)
\end{equation} 
The \textit{exponential map} at $p$
\begin{subequations}\label{eq:exp-Riemannian}
\begin{align}
\exp_{ \mc{M}, p} \colon V_{p} &\to \mc{M},\qquad
v\mapsto \exp_{\mc{M},}p(v) \coloneqq \gamma_{v}(1)
\intertext{
is defined on 
}
V_{p} &= \{v \in T_{p}\mc{M} \colon \gamma_{v}\;\text{is defined on}\;[0,1]\}
\end{align}
\end{subequations}
in terms of the \textit{geodesic} $\gamma_{v}(t)$ through $p=\gamma_{v}(0)$ with velocity $v =\dot\gamma_{v}(0)$.

The \textit{weighted Riemannian mean} \cite[Def.~6.9.1]{Jost:2017aa} of a collection $p_{1},\dotsc,p_{n} \in \mc{M}$ of points with respect to weights $w = (w_{1},\dotsc,w_{n}) \in \Delta_{n}$ is the point $q \in \mc{M}$ satisfying 
\begin{equation}\label{eq:Jw-Riemannian}
J_{w}(q) = \inf_{p \in \mc{M}} J_{w}(p),\qquad
J_{w}(p) = \frac{1}{2} \sum_{i \in [n]} w_{i} d_{g}^{2}(p_{i},p),
\end{equation}
where $d_{g}(q,p)$ denotes the \textit{Riemannian distance}, i.e.~the infimum of the length of all smooth paths connecting $q$ and $p$ on $\mc{M}$. We have \cite[Lemma 6.9.4]{Jost:2017aa}
\begin{subequations}\label{eq:grad-J-general}
\begin{align}
\ggrad_{p} J_{w} 
&= -\sum_{i \in [n]} w_{i} \exp_{\mc{M},p}^{-1}(p_{i}) \in T_{p}\mc{M}
\intertext{
and hence the optimality condition for $q$
}
\sum_{i \in [n]} w_{i}\exp_{\mc{M},q}^{-1}(p_{i}) &= 0. \label{eq:grad-J-general-opt}
\end{align}
\end{subequations}
This equation is typically solved by the \textit{mean shift} (fixed point) \textit{iteration}
\begin{equation}\label{eq:Rmean-fp-iteration}
q^{(t+1)} = \exp_{\mc{M},q^{(t)}}\Big(\sum_{i \in [n]} w_{i} \exp_{\mc{M},q^{(t)}}^{-1}(p_{i})\Big),\quad t=1,2,\dotsc
\end{equation}
with a suitable initialization $q^{(0)}$.

\subsection{Divergence Functions}
\label{sec:Divergence-Functions}
\textit{Bregman divergences} are distance-like functions of the form
\begin{equation}\label{eq:D-Bregman}
D_{\phi} \colon \dom\phi \times \intr(\dom\phi) \to\R_{+},
\qquad
D_{\phi}(x,y) = \phi(x)-\phi(y)-\la\nabla\phi(y),x-y\ra, 
\end{equation}
induced by smooth convex functions $\phi \colon \R^{d} \to \R$ of Legendre type \cite{Censor:1997aa,Bauschke:1997aa}.
The second argument of $D_{\phi}$ is admissible only in the interior of the domain of $\phi$, where $\phi$ is continuous differentiable with finite gradient $\nabla \phi$.
Divergences $D_{\phi}$ satisfy
\begin{subequations}\label{eq:Bregman-properties}
\begin{align}\label{eq:Bregman-properties-a}
&D_{\phi} \geq 0 \quad\text{and}\quad D_{\phi}(x,y)=0\;\gdw\;x=y,
\\ \label{eq:Bregman-properties-b}
&\nabla_{x}^{2} D_{\phi}(x,y) \succ 0,\quad\forall x \in \intr(\dom\phi).
\end{align}
\end{subequations}
The former property shows that $D_{\phi}$ behaves like a distance, but symmetry $D_{\phi}(x,y)=D_{\phi}(y,x)$ is not required and generally does not hold. The second property \eqref{eq:Bregman-properties-b}, i.\,e., the Hessian is positive definite, shows that $D_{\phi}$ can be used to define a Riemannian metric in order to turn an open subset of a Euclidean space into a manifold.
 
More generally, given a $d$-dimensional Riemannian manifold $(\mc{M},g)$, a function $D_{\phi} \colon \mc{M} \times \mc{M} \to \R_{+}$ is a proper divergence function 
defined on $\mc{M}$ if, for any chart $U \subset \mc{M}$ with local coordinates $x \colon U \to \R^{d}$ and $p, q \in U$, the function
\begin{subequations}\label{eq:DM-properties}
\begin{align}
\widetilde D_{\phi}\big(x(p),x(q)\big) &= D_{\phi}(p,q)
\intertext{
satisfies \eqref{eq:Bregman-properties-a} and recovers the positive definite metric tensor by
}
D_{\phi}(p,q) &\approx \frac{1}{2} \sum_{i,j \in [d]} g_{ij}(p) z_{i} z_{j},
\end{align}
\end{subequations}
for $z = x(q)-x(p)$ and small $\|z\|$. 

Two further facts are relevant to the present paper. First, the function $D_{\phi}(p,q)=\frac{1}{2} d_{g}(p,q)^{2}$ defines a \textit{canonical} divergence function on a Riemannian manifold $(\mc{M},g)$ in terms of the squared Riemannian distance $d_{g}^{2}$. Secondly, alternative divergence functions $D_{\phi}$ satisfying \eqref{eq:DM-properties} are required in many applications, that serve as surrogate functions in \eqref{eq:Jw-Riemannian} for the squared Riemannian distance $d_{g}^{2}(p_{i},p)$ and are easier to evaluate computationally. Concrete divergence functions in connection with unsupervised label learning are studied in Section~\ref{sec:Feature-Manifolds}.

Likewise, in information geometry, the Riemannian (Levi-Civita) connection is replaced by another affine connection in order to define a divergence function through affine geodesics and corresponding squared distances. We refer to \cite[section 3.4]{Amari:2000aa}, \cite{Amari:2010aa} and \cite[section 4.4]{Ay:2017aa} for background and further details. A concrete application is provided by the assignment manifold (Section \ref{sec:Assignment-Supervised})  and corresponding concepts defining the assignment flow in Section \ref{sec:Assignment-Supervised}. 

\section{Basic Clustering}
\label{sec:Preliminaries-Clustering}

\noindent
We briefly summarize in this section the basic iterative schemes 
\begin{itemize}
\item
soft-$k$-means clustering in Euclidean spaces (Section \ref{sec:Euclidean-k-Means}),
\item 
clustering using mixture distributions, divergence functions and the EM-algorithm (Section \ref{eq:Divergences-EM}), and 
\item
greedy-based clustering in metric spaces (Section \ref{sec:Metric-Clustering}). 
\end{itemize}
The first two approaches will be generalized to \textit{manifold-valued} data (features) in Section \ref{sec:Manifold-Clustering} and coupled with the assignment flow for spatial regularization in Section \ref{sec:Assignment-Unsupervised}. 

Greedy-based $k$-center clustering applies to any metric space, in particular to manifolds with the Riemannian distance or a suitable divergence as a surrogate distance function. The method has linear complexity and comes along with a performance guarantee. Hence, this method is suited for fast data selection in a preprocessing step, to obtain an overcomplete codebook (set of prototypes) as initialization for manifold-valued clustering, which subsequently optimizes and sparsifies this codebook in a computationally more expensive way.

\subsection{Euclidean Soft-$k$-Means Clustering}
\label{sec:Euclidean-k-Means}

The content of this paragraph can be found in numerous papers and textbooks. We merely refer to the survey \cite{Teboulle:2007aa} and to the bibliography therein. 

Given data vectors $x^{1},\dotsc,x^{|I|} \in \R^{d}$, we consider the task of determining prototype vectors 
\begin{equation}\label{eq:M-Euclidean}
M = \{m^{1},\dotsc,m^{|J|}\} \subset \R^{d}
\end{equation}
by minimizing the $k$-means criterion\footnote{%
The symbol `$k$' is commonly used in the literature. We prefer  in this paper, however, the more specific symbol $J$ as index set for prototypes and use $k$ (like $i, j$ etc.) as free index.%
} 
\begin{equation}\label{eq:JM-Euclidean}
E(M) = \sum_{i \in I} \min_{j \in J}\|x^{i}-m^{j}\|^{2}
= \sum_{i \in I} \vecmin(D_{i}(M)\big),
\end{equation} 
where
\begin{equation}\label{eq:diM-Euclidean}
D_{i}(M)=\big(D_{i1}(M),\dotsc,D_{i|J|}(M)\big)
= \big(\|x^{i}-m^{1}\|^{2},\dotsc,\|x^{i}-m^{|J|}\|^{2}\big)
\end{equation}
and
\begin{equation}\label{eq:vecmin}
\vecmin(z) = \min_{j \in [d]}\{z_{1},\dotsc,z_{d}\},\quad z \in \R^{d},\quad d \in \N.
\end{equation}
\textit{Soft}-$k$-means  is based on the \textit{smoothed} objective 
\begin{equation}\label{eq:def-Jeps-Euclidean}
E_{\veps}(M) = -\veps \sum_{i \in I} \log\Big(
\sum_{j \in J} \exp\big(-\frac{\|x^{i}-m^{j}\|^{2}}{\veps}\big)\Big), \quad \veps > 0
\end{equation}
which results from approximating the inner minimization problem of evaluating $E(M)$ using the log-exponential function \cite[p.~27]{Rockafellar:2009aa} with smoothing parameter $\veps$. Similar to the basic $k$-means algorithm, \textit{soft}-$k$-means clustering solves the stationarity conditions 
\begin{equation}\label{eq:dJmj-Euclidean}
\nabla_{m^{j}} E_{\veps}(M) = 0,\quad j \in J
\end{equation}
by fixed point iteration in terms of iteratively computing the \textbf{soft-assignments}
\begin{subequations}\label{eq:soft-k-means-iteration}
\begin{gather}\label{eq:soft-k-means-p}
p_{\veps,j}^{i}(M) = \frac{\exp\big(-D_{ij}(M)/\veps\big)}
  {\sum_{k \in J} \exp\big(-D_{ik}(M)/\veps\big)},\quad
q_{\veps,i}^{j}(M) = \frac{p_{\veps,j}^{i}(M)}{\sum_{k \in I} p_{\veps,j}^{k}(M)},\quad
i \in I,\; j \in J
\intertext{
with the so-called \textbf{mean shifts}
} \label{eq:mean-shift-Euclidean}
m^{j} = \sum_{i \in I} q_{\veps,i}^{j}(M) x^{i},\quad j \in J.
\end{gather}
\end{subequations} 
The distributions 
\begin{equation}\label{eq:def-p-Euclidean}
p^{i}_{\veps}(M) \in \Delta_{|J|},\quad i \in I
\end{equation}
given by \eqref{eq:soft-k-means-p} represent the soft-assignments $p^{i}_{\veps,j}(M)$ of each data point $x^{i},\,i \in I$ to each prototype $m^{j},\,j \in J$, whereas the distributions
\begin{equation}\label{eq:def-q-Euclidean}
q_{\veps}^{j}(M) \in \Delta_{|I|},\quad j \in J
\end{equation}
determine the convex combinations of data points that determine each prototype $m^{j}$ by the mean shift \eqref{eq:mean-shift-Euclidean}. Iterating the two steps \eqref{eq:soft-k-means-iteration} evolves the prototypes $M$ until they reach a local minimum of the objective \eqref{eq:def-Jeps-Euclidean}.

\subsection{Divergence Functions and EM-Iteration}
\label{eq:Divergences-EM}

An alternative and widely applied approach to clustering utilizes class-conditional distributions 
$p(x;\theta_{j}),\; j \in J$ 
and a corresponding mixture distribution
\begin{equation}\label{eq:mixture-model}
p(x;\Gamma) = \sum_{j \in J} \pi_{j} p(x;\theta_{j})
\end{equation}
as data model, with parameters
\begin{equation}
\Gamma = (\theta,\pi),\qquad
\theta=(\theta_{1},\dotsc,\theta_{|J|}),\qquad
\pi = (\pi_{1},\dotsc,\pi_{|J|})^{\T} \in \mc{S},
\end{equation}
with the relative interior of the probability simplex 
\begin{equation}\label{eq:def-mcS0}
\mc{S} \coloneqq \rint \Delta_{|J|} = \{p \in \R^{|J|} \colon p_{j}>0,\; j \in J,\; \la\eins,p \ra = 1\}.
\end{equation}
Clustering amounts to estimate the parameters $\Gamma$. Since the log-likelihood function corresponding to \eqref{eq:mixture-model} is usually involved, maximizing a lower bound through the EM-iteration (EM: expectation-maximization) is the method of choice,
\begin{subequations}
\begin{align}
p(j|x^{i};\Gamma^{(t)}) &= \frac{\pi_{j}^{(t)} p(x^{i};\theta_{j}^{(t)})}{\sum_{l \in J}\pi_{l}^{(t)} p(x^{i};\theta_{l}^{(t)})},\qquad j \in J \qquad\qquad(\textbf{E-step, soft-assignment}) \\[0.5cm]
&
\begin{aligned}
\mathllap{\pi_{j}^{(t+1)}} &= \frac{1}{|J|} \sum_{i \in I} p(j|x^{i};\Gamma^{(t)})
\\
\mathllap{\theta_{j}^{(t+1)}} &= \argmax_{\theta_{j}} \sum_{i \in I} p(j|x^{i};\Gamma^{(t)}) \log p(x^{i};\theta_{j})
\end{aligned}
,\qquad j \in J. \qquad\qquad (\textbf{M-step})
\end{align}
\end{subequations}
for some initialization $\Gamma^{(0)}$. 
We refer to \cite{McLachlan:2000aa} for background and further details.

Banerjee et al.~\cite{Banerjee:2005aa} studied the case where the class-conditional distributions $p(x;\theta_{j})$ of \eqref{eq:mixture-model} belong to an exponential family of distributions \cite{Barndorff-Nielsen:1978aa} and, in particular, their representation in terms of a Bregman divergence function $D_{\phi}$. Then the resulting data model \eqref{eq:mixture-model} reads
\begin{equation}\label{eq:mixture-Bregman}
p(x;\Gamma) = \sum_{j \in J} \pi_{j} \exp\big(-D_{\phi}(f(x),\eta_{j})\big) b_{\phi}(x),
\end{equation}
where $f$ denotes a sufficient statistics regarded as a feature vector, the factor $b_{\phi}$ accounts for normalization and $\eta_{j} = \nabla \psi(\theta_{j})$ is determined by $\theta_{j}$ through conjugation of the convex log-partition function $\psi(\theta_{j})= \log\int_{\mc{X}} p(x;\theta_{j})\dd{x}$. The corresponding EM-updates read
\begin{subequations}\label{eq:EM-Bregman}
\begin{align}
p(j|x^{i};\Gamma^{(t)}) &= \frac{\pi_{j}^{(t)} \exp\big(-D_{\phi}(f(x^{i}),\eta_{j}^{(t)})\big)}{\sum_{l \in J}\pi_{l}^{(t)} \exp\big(-D_{\phi}(f(x^{i}),\eta_{l}^{(t)})\big)},\qquad j \in J \qquad(\textbf{E-step, soft-assignment})
\\[0.5cm]
&
\begin{aligned}
\mathllap{\pi_{j}^{(t+1)}} &= \frac{1}{|J|} \sum_{i \in I} p(j|x^{i};\Gamma^{(t)})
\\ \label{eq:eta-update-EM}
\mathllap{\eta_{j}^{(t+1)}} &= \argmin_{\eta_{j}} \sum_{i \in I} \nu_{j|i}(\Gamma^{(t)}) D_{\phi}\big(f(x^{i}),\eta_{j}^{(t)}\big)
\end{aligned}
,\qquad j \in J. \qquad\qquad (\textbf{M-step})
\\[0.5cm]
\nu_{j|i}(\Gamma^{(t)}) &= \frac{p(j|x^{i};\Gamma^{(t)})}{\sum_{k \in I} p(j|x^{k};\Gamma^{(t)})}.
\end{align}
\end{subequations}
Moreover, since the Bregman divergence $D_{\phi}$ is induced by a convex function $\phi$ of Legendre type, the parameters $\eta_{j},\, j \in j$ can be updated by the \textbf{mean-shifts}
\begin{equation}\label{eq:eta-update-explicit-EM}
\eta_{j}^{(t+1)} = \sum_{i \in I} \nu_{j|i}(\Gamma^{(t)}) f(x^{i}),\quad j \in J.
\end{equation}
We exploit the above connection to divergence functions in Sections \ref{sec:manifold-EM} and \ref{sec:regularized-EM-manifold}.

\subsection{Greedy-Based $k$-Center Clustering in Metric Spaces}
\label{sec:Metric-Clustering}

\begin{figure}[ht]
\centerline{
\includegraphics[width=0.25\textwidth]{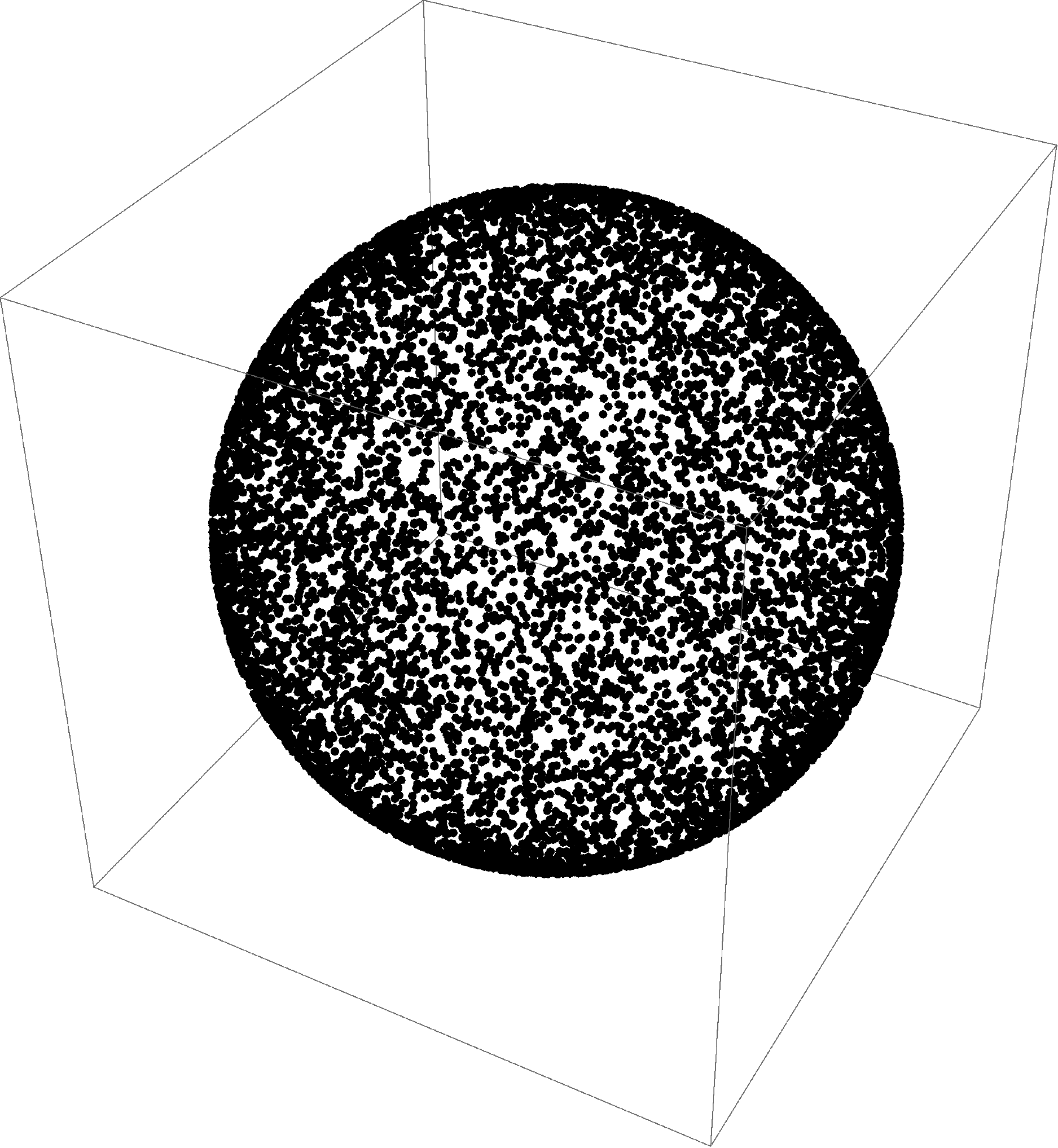}
\hspace{0.05\textwidth}
\includegraphics[width=0.25\textwidth]{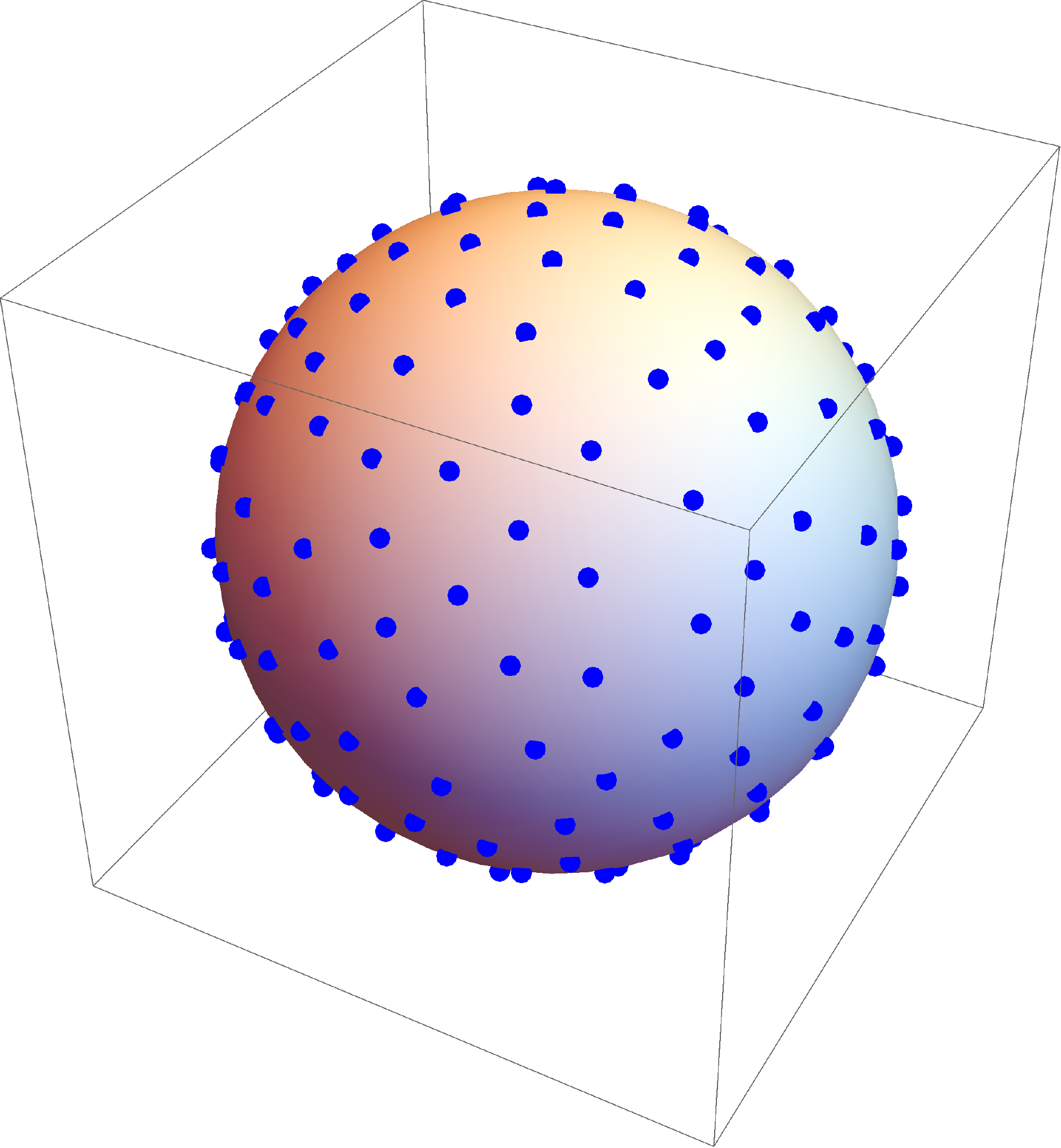}
}
\caption{
\textbf{Approximation of the $k$-center clustering objective \eqref{eq:metric-clustering-objective}.} \textsc{left:} 10.000 points on the sphere regarded as manifold equipped with the cosine distance. \textsc{right:} 200 prototypes determined with linear runtime complexity by metric clustering are \textit{almost uniformly located} in the data set, which qualifies them for unbiased initializations of computationally more involved nonlinear prototype evolutions. This works in any metric space, in particular on feature manifolds using the Riemannian distance or computationally less expensive divergence functions.
}
\label{fig:SphereClustering}
\end{figure}

We adopt a simple greedy algorithm from \cite{Har-Peled:2011aa} as a \textit{preprocessing step for data reduction}, due to the following properties:
\begin{itemize}
\item It works in any \textit{metric space} $(X,d_{X})$,
\item it has \textit{linear} complexity $\mc{O}(|J| |I|)$ with respect to the problem size $|I|$ which can be  large,
\item it comes along with a \textit{performance guarantee}.
\end{itemize}
The task of $k$-center clustering is as follows.
Given data points 
\begin{equation}
X_{I}=\{x^{1},\dotsc,x^{|I|}\} \subset X,
\end{equation}
the objective is to determine a subset 
\begin{equation}
M=\{m^{1},\dotsc,m^{|J|}\} \subset X_{I}
\end{equation}
that solves the combinatorially hard optimization problem
\begin{equation}\label{eq:metric-clustering-objective}
E_{\infty}^{\ast} = \min_{M \subset X_{I}, |M|=|J|} E_{\infty}(M),\qquad 
E_{\infty}(M) = \max_{x \in X_{I}} d_{X}(x,M),
\end{equation}
where $d_{X}(x,M) = \min_{m \in M} d_{X}(x,m)$. 
This problem is approximated by a greedy iteration: 
Starting with a randomly chosen point $m^1 \in X_I$, the remaining $|J|-1$ points $m^{2},\dotsc,m^{|J|}$ are selected by choosing $m^{k+1} \in X_I$ with the largest distance to the points $m^1, \dots, m^k$. 
This greedy strategy possesses the following performance guarantee: 
The resulting set $M$ is a $2$-approximation of the optimum \eqref{eq:metric-clustering-objective}, i.\,e. $E_{\infty}(M) \leq 2 E_{\infty}^*$. We refer to \cite[Thm.~4.3]{Har-Peled:2011aa} for the proof.
As a consequence, the subset $M$ of $|J|$ points is almost uniformly distributed in $X_{I}$ as measured by the metric $d_{X}$. 
Figure \ref{fig:SphereClustering} provides an illustration.

We note that the performance guarantee follows by the triangle inequality, while the complexity is independent of the properties of a metric space.
We will later apply this algorithm in the context of clustering on a Riemannian manifold $\mc{M}$, where we have given a smooth (symmetric) divergence function $D$ on $\mc{M}$. 
Most of these divergence functions are squared distances, so that the performance guarantee also holds in this case. One exception might be the rotation-invariant dissimilarity \eqref{eq:stein-rotation-inv}. Nevertheless, the greedy $k$-center clustering is applicable but without having a performance guarantee.
We will use greedy $k$-center clustering as preprocessing in order to get an overcomplete set of labels as initial labels for the clustering approaches described in the next section.

%

\section{Coupling Clustering on Manifolds and Spatially Regularized Assignment}
\label{sec:Unsupervised-Manifold-Assignment-Flow}

We reformulate in Section \ref{sec:Manifold-Clustering} the iterative schemes of Sections \ref{sec:Euclidean-k-Means} and \ref{eq:Divergences-EM} in order to cope with \textit{manifold-valued data}. Both schemes will be coupled in Section \ref{sec:Assignment-Unsupervised} with the assignment flow that is presented in Section \ref{sec:Assignment-Supervised}. This results in two novel schemes for spatially regularized label (prototype) learning from manifold-valued data. Finally, we define in Section \ref{sec:coupled-flow-final} the \textit{unsupervised assignment flow} as smooth interpolation of the flows corresponding to both schemes, depending on a single interpolation parameter.

\subsection{Manifold-Valued Clustering}
\label{sec:Manifold-Clustering}
We generalize the basic iterative clustering schemes of Sections \ref{sec:Euclidean-k-Means} and \ref{eq:Divergences-EM} to manifold-valued data.

\subsubsection{Manifold-Valued Soft-$k$-Means Iteration}\label{sec:soft-k-manifold}
Let $(\mc{M},g)$ be a smooth Riemannian manifold and let
\begin{equation}\label{eq:data-mcM}
\{z^{1},\dotsc,z^{|I|}\} \subset \mc{M}
\end{equation}
be given data. 
We assume a smooth divergence function to be given (cf.~Section \ref{sec:Divergence-Functions})
\begin{equation}\label{eq:def-divergence}
D \colon \mc{M} \times \mc{M} \to \R,\qquad
(x,y)\mapsto D(x,y)
\end{equation}
that replaces the Riemannian distance $d_{g}$ in order to compute Riemannian means more efficiently or even in closed form. We just use the symbol ``$D$'' and omit the subscript of \eqref{eq:D-Bregman}, because what follows applies to various scenarios and to any corresponding concrete divergence function $D$. Examples are provided in Section \ref{sec:Feature-Manifolds}.

We consider the task to determine a set of prototypes
\begin{equation}\label{eq:labels-mcM}
M = \{m^{1},\dotsc,m^{|J|}\} \subset \mc{M}
\end{equation}
by minimizing an objective function analogous to the soft-$k$-means objective \eqref{eq:def-Jeps-Euclidean},
\begin{equation}\label{eq:geom-soft-k-means-objective}
E_{\veps}(M) \coloneqq E_{\veps}(m^{1},\dotsc,m^{|J|}) = -\veps \sum_{i \in I} \log\Bigg(\sum_{j \in J} \exp\Big(-\frac{D(z^{i},m^{j})}{\veps}\Big)\Bigg),\quad \veps > 0.
\end{equation}
We next generalize the conditions \eqref{eq:dJmj-Euclidean}.  
Let $d_{j}E_{\veps}(M)$ denote the differential of the function \\
$m^{j}\mapsto E_{\veps}(m^{1},\dotsc,m^{j},\dotsc,m^{|J|})$. Then
\begin{equation}\label{eq:djJM-manifold}
d_{j}E_{\veps}(M) = \sum_{i \in I} \underbrace{\frac{\exp\big(-\frac{D(z^{i},m^{j})}{\veps}\big)}{\sum_{l \in J} \exp\big(-\frac{D(z^{i},m^{l})}{\veps}\big)}}_{\coloneqq p^{i}_{\veps,j}(M)} d_{j}D(z^{i},m^{j})
= \sum_{i \in I} p^{i}_{\veps,j}(M) d_{j}D(z^{i},m^{j}),\quad
j \in J,
\end{equation}
where the \textbf{assignment probability vectors} $p^{i}_{\veps}(M) \in \Delta_{|J|}$ play the same role as in eqns.~\eqref{eq:soft-k-means-p} and \eqref{eq:def-p-Euclidean}. They can be interpreted as \textit{weight functions} depending on the prototypes $M$: setting temporarily $w_{i} = p^{i}_{\veps,j}(M)$, $i \in I$, implies that eq.~\eqref{eq:djJM-manifold} has the same structure as the  equation on the right of \eqref{eq:Jw-Riemannian} after applying the differential on both sides, where we take into account that divergence functions $D(\cdot,\cdot)$ behave like squared distances (Section \ref{sec:Divergence-Functions}). 
Applying formula \eqref{eq:def-grad-4}, we obtain the gradients and optimality conditions
\begin{equation}\label{eq:grad-dJj-manifold}
(\ggrad E_{\veps})_{j}(M) = \widehat g^{-1}\big(d_{j}E_{\veps}(M)\big) = \sum_{i \in I} p^{i}_{\veps,j}(M) \widehat g^{-1}\big(d_{j}D(z^{i},m^{j})\big) = 0,\quad j \in J.
\end{equation}
Comparing with \eqref{eq:dJmj-Euclidean} shows that, in the Euclidean case, the mean shift operation \eqref{eq:mean-shift-Euclidean} is defined by \textit{normalized} weights $q^{j}_{\veps,i}(M)$ due to \eqref{eq:soft-k-means-p}, conforming to the much more general situation \eqref{eq:grad-J-general}. While normalization in \eqref{eq:soft-k-means-p} is a \textit{consequence} of the squared Euclidean distance of the objective \eqref{eq:def-Jeps-Euclidean}, this may or may not happen in \eqref{eq:grad-dJj-manifold}, depending on the particular manifold $\mc{M}$, metric $g$ and divergence function $D$ at hand. Because subdividing each optimality condition \eqref{eq:dJmj-Euclidean} by the corresponding normalization factor in \eqref{eq:soft-k-means-iteration} does not change the condition, however, and because mean-shift on manifolds is performed with normalized weights, we define
\begin{equation}
p^{i}_{\veps,j}(M)
= \frac{\exp\big(-\frac{D(z^{i},m^{j})}{\veps}\big)}{\sum_{l \in J} \exp\big(-\frac{D(z^{i},m^{l})}{\veps}\big)},\qquad
q^{j}_{\veps,i}(M)
= \frac{p^{i}_{\veps,j}(M)}{\sum_{k \in I} p^{k}_{\veps,j}(M)},\quad i \in I,\; j \in J
\end{equation}
and in turn the \textbf{mean shift} (fixed point) \textbf{iteration}
\begin{equation}\label{eq:fp-iter-M-general}
(m^{j})^{(t+1)} = \exp_{\mc{M},(m^{j})^{(t)}}\Big(\sum_{i \in I}
q^{j}_{\veps,i}(M^{(t)}) \widehat g^{-1}\big(d_{j}D((m^{j})^{(t)},z^{i})\big)\Big),\quad j \in J
\end{equation}
analogous to \eqref{eq:Rmean-fp-iteration}. Section \ref{sec:Feature-Manifolds} provides concrete examples for divergence functions on manifolds.

\subsubsection{Manifold-Valued EM-Iteration}
\label{sec:manifold-EM}
We consider again the situation \eqref{eq:data-mcM}--\eqref{eq:labels-mcM} and adopt the clustering approach of Section \ref{eq:Divergences-EM}. Iteration \eqref{eq:EM-Bregman} generalizes to
\begin{subequations}\label{eq:EM-manifold}
\begin{align}\label{eq:pj|i-EM}
p(j|z^{i};M^{(t)}) &= \frac{\pi_{j}^{(t)} \exp\big(-D(z^{i},(m^{j})^{(t)})\big)}{\sum_{l \in J}\pi_{l}^{(t)} \exp\big(-D(z^{i},(m^{l})^{(t)})\big)},\qquad j \in J \qquad(\textbf{E-step, soft-assignment})
\\[0.5cm]
&
\begin{aligned}
\mathllap{\pi_{j}^{(t+1)}} &= \frac{1}{|J|} \sum_{i \in I} p(j|z^{i};M^{(t)})
\\ \label{eq:mj-update-EM-manifold}
\mathllap{(m^{j})^{(t+1)}} &= \argmin_{m^{j}}~\sum_{i \in I} \nu_{j|i}(M^{(t)}) D\big(z^{i},(m^{j})^{(t)}\big)
\end{aligned}
,\qquad j \in J. \qquad\qquad (\textbf{M-step})
\\[0.5cm]
\nu_{j|i}(M^{(t)}) &= \frac{p(j|z^{i};M^{(t)})}{\sum_{k \in I} p(j|z^{k};M^{(t)})}.
\end{align}
\end{subequations}
Note that we apparently ignore here the connection to class-conditional distributions $p(x;\theta_{j})$ of the exponential family that formed the basis for the EM-iteration \eqref{eq:EM-Bregman}. This is not the case, however. Indeed, optimization problem \eqref{eq:mj-update-EM-manifold} which determines each prototype $m^{j}$ by minimizing the \textit{expected value} of a squared distance-like function conforms to the updates \eqref{eq:eta-update-EM} and \eqref{eq:eta-update-explicit-EM} of the \textit{expectation parameter} $\eta_{j} = \nabla\psi(\theta_{j}) = \EE_{\theta_{j}}[f_{i}]$, where the expectation is with respect to $p(x;\theta_{j})$ and the sufficient statistics $f_{i}(x)$.

In order to solve problem \eqref{eq:mj-update-EM-manifold}, we proceed analogously to \eqref{eq:grad-dJj-manifold}. Examples with concrete choices of $D(\cdot,\cdot)$ are worked out in Section \ref{sec:Feature-Manifolds}.

\subsection{Supervised Assignment Flow}
\label{sec:Assignment-Supervised}
In this section, we review the assignment flow introduced in \cite{Astrom:2017ac}. An overview of more recent work can be found in \cite{Schnorr:2019aa}. For the (information) geometric aspects of the probability simplex, we refer to \cite[Section 2.5]{Amari:2000aa}.

Let data be given by \eqref{eq:data-mcM} together with  \textit{fixed} labels (prototypes) \eqref{eq:labels-mcM}. The index set $I$ corresponds to pixel locations $i \in I$ and extracted features $z^{i},\, i \in I$, whereas the index set $J$ enumerates the labels (class representatives, prototypes) $m^{j},\, j \in J$. After fixing a suitable divergence function \eqref{eq:def-divergence}, the \textbf{distance vectors}
\begin{equation}\label{eq:def-Di}
D_{i}(M)
= \big(D(z^{i},m^{1}),\dotsc,D(z^{i},m^{|J|})\big) \in \R^{|J|},\quad i \in I
\end{equation}
are defined. The approach \cite{Astrom:2017ac} is based on the relatively open \textit{probability simplex} of strictly positive vectors 
\begin{equation}\label{eq:def-mcS}
\mc{S} = \{p \in \R^{|J|} \colon p_{j}>0,\; j \in J,\; \la\eins,p \ra = 1\}
\end{equation}
with the uniform distribution as barycenter,
\begin{equation}\label{eq:def-barycenter}
\eins_{\mc{S}} \coloneqq \frac{1}{|J|} \eins_{|J|},
\qquad\qquad(\textbf{barycenter} \;\text{of $\mc{S}$})
\end{equation}
that becomes a Riemannian manifold when equipped with the \textit{Fisher-Rao metric}
\begin{equation}
g_{p}(u,v) = \sum_{j \in J} \frac{u_{j} v_{j}}{p_{j}},\quad u,v, \in T_{0},\quad p \in \mc{S},
\end{equation}
where $T_{0}$ denotes the tangent space
\begin{equation}
T_{0} \coloneqq T_{\eins_{\mc{S}}}\mc{S} = \{v \in \R^{|J|} \colon \la\eins,v\ra=0\},\quad p \in \mc{S}
\end{equation}
that we will work with throughout this paper, in lieu of the tangent spaces $T_{p}\mc{S} = \{u = \frac{v}{p} \colon v \in T_{0} \}$, $p \in \mc{S}$ that are equivalent up to the normalization. We denote by
\begin{equation}\label{eq:def-Pi-p}
R_{p} \colon \R^{|J|} \to T_{0},\qquad
d \mapsto R_{p}(d) \coloneqq (\Diag(p) - p p^{\T}) d
= p (d - \la p, d \ra \eins),\quad p \in \mc{S}
\end{equation}
a family of linear mappings onto $T_{0}$ parameterized by $p$. 

Adopting the $\alpha$-connection with $\alpha=1$ from information geometry as introduced by Amari and Chentsov \cite[Section 2.3]{Amari:2000aa}, affine geodesics and a corresponding \textit{exponential map} are given by
\begin{equation}\label{eq:def-exp-S}
\Exp \colon \mc{S} \times T_{0} \to \mc{S},\qquad
(p,v) \mapsto \Exp_{p}(v) \coloneqq \frac{p e^{\frac{v}{p}}}{\la p, e^{\frac{v}{p}} \ra} .
\end{equation}
These geodesics are not length minimizing unlike the geodesics induced by the Riemannian (Levi-Civita) connection, but they closely approximate them  \cite[Prop.~3]{Astrom:2017ac} and are computationally more convenient to work with. In particular, unlike exponential maps in general (cf.~\eqref{eq:exp-Riemannian}), the map $\Exp_{p}$ is defined on the entire space $T_{0}$ and has the inverse \cite[Appendix]{Astrom:2017ac}
\begin{equation}\label{eq:def-invexp}
\Exp^{-1} \colon \mc{S} \times \mc{S} \to T_{0}, 
\qquad\qquad
(p,q) \mapsto \Exp_{p}^{-1}(q) = R_{p}\log\frac{q}{p},
\end{equation}
with $R_{p}$ given by \eqref{eq:def-Pi-p}. The composition of \eqref{eq:def-exp-S} and \eqref{eq:def-Pi-p} defines the map\footnote{Note that the symbol $\exp_{p}$ does not contain the subscript $\mc{S}$ in order to distinguish it from the definition \eqref{eq:exp-Riemannian} for general manifolds $\mc{M}$}
\begin{equation}\label{eq:def-exp-1}
\exp_{p} 
\coloneqq \Exp_{p} \circ R_{p} \colon \R^{|J|} = T_{0} \oplus \R\eins\to \mc{S}, 
\qquad\qquad
z \mapsto \frac{p e^{z}}{\la p, e^{z}\ra},\qquad p \in \mc{S}.
\end{equation}
These mappings apply to 
\textbf{assignment vectors} 
\begin{equation}\label{eq:def-Wi}
W_{i} \in \mc{S},\quad i \in I,
\end{equation}
associated with each pixel $i \in I$ that represent the a posteriori probabilities
\begin{equation}\label{eq:Wij-Pr}
W_{ij} = \Pr(j|z^{i}),\quad i \in I,\; j \in J.
\end{equation}
The assignment vectors form the row vectors of the \textbf{assignment matrix}
\begin{equation}\label{eq:def-assignment-matrix}
W = \bpm \vdots \\ W_{i}^{\T} \\ \vdots \epm = \bpm \dotsb & W^{j} & \dotsb \epm \in \mc{W} \subset \R_{++}^{|I| \times |J|},
\end{equation}
whose column vectors are denoted by $W^{j},\, j \in J$. Due to \eqref{eq:def-Wi}, $W \in \mc{W}$ is 
regarded as point on the
\begin{equation}\label{eq:def-mcW}
\mc{W} \coloneqq \mc{S} \times \dotsb \times \mc{S} \qquad(|I|\; \text{times})\qquad\qquad(\textbf{assignment manifold})
\end{equation}
with tangent space
\begin{equation}\label{eq:def-T0-product}
\mc{T}_{0} \coloneqq T_{0} \times \dotsb \times T_{0} \qquad(|I|\; \text{times})
\end{equation}
and the corresponding mappings
\begin{subequations}\label{eq:assignment-flow-maps}
\begin{align}
\eins_{\mc{W}} 
&\coloneqq (\eins_{\mc{S}},\dotsc,\eins_{\mc{S}}) \in \mc{W}
&& (\textbf{barycenter})
\\ 
R_{W}(Z) 
&\coloneqq \big(R_{W_{1}}(Z_{1}),\dotsc,R_{W_{|I|}}(Z_{|I|})\big) \in \mc{T}_{0}, &
W &\in \mc{W},\quad Z \in \R^{|I|\times|J|}
\\ \label{eq:exp-e-W}
\Exp_{W}(V) 
&\coloneqq \big(\Exp_{W_{1}}(V_{1}),\dotsc,\Exp_{W_{|I|}}(V_{|I|})\big) \in \mc{W}, &
W &\in \mc{W},\quad V \in \mc{T}_{0}
\end{align}
\end{subequations}
and $\exp_{W}, \Exp^{-1}_{W}$ similarly defined based on \eqref{eq:def-exp-1}, \eqref{eq:def-invexp}.

We assume neighborhoods 
\begin{equation}\label{eq:def-Ni}
\mc{N}_{i} = \{k \in I \colon ik \in \mc{E}\} \cup \{ i \},\quad i \in I
\end{equation}
to be defined around each pixel $i \in I$, formally given by a graph $G=(I,\mc{E})$ with pixel indices $I$ as vertex set and edges $\mc{E}$ defining \eqref{eq:def-Ni}. We associate with each neighborhood $\mc{N}_{i}$ weights $\{w_{ik} \colon k \in \mc{N}_{i}\}$ satisfying
\begin{equation}\label{eq:def-weights}
w_{ik} > 0,\qquad
\sum_{k \in \mc{N}_{i}} w_{ik} = 1,\qquad \forall i \in I.
\end{equation}
These weights parameterize the regularization property of the assignment flow and are assumed to be given. We refer to \cite{Huhnerbein:2019aa} for an approach to learn these parameters from data. \eqref{eq:def-Ni} and \eqref{eq:def-weights} define the \textit{geometric mean} of assignment vectors \cite[Lemma 5]{Astrom:2017ac}
\begin{equation}\label{eq:def-geom-mean}
\mc{G}^{w}_{i}(W) \coloneqq \Exp_{W_{i}} \Big(\sum_{k \in \mc{N}_{i}} w_{ik} \Exp_{W_{i}}^{-1}(W_{k})\Big)
= \exp_{W_{i}}\Big(\log\frac{\prod_{k \in \mc{N}_{i}} W_{k}^{w_{ik}}}{W_{i}}\Big),\qquad i \in I,
\end{equation}
which defines -- specifically for the assignment manifold \eqref{eq:def-mcW} -- a closed-form solution to the general equation \eqref{eq:grad-J-general-opt} that can be computed efficiently.

\vspace{0.2cm}
Using this setting, the assignment flow accomplishes image labeling as follows. Based on \eqref{eq:def-Di} 
\begin{equation}\label{eq:def-Di2}
D = (D_{1},\dotsc,D_{|I|}) \in \R^{|I|\times|J|}
\qquad\qquad\qquad\qquad
(\textbf{distance vectors})
\end{equation}
are defined and mapped to
\begin{subequations}\label{eq:def-LW}
\begin{align}
L(W) &= \exp_{W}(D) \in \mc{W},
\qquad\qquad\qquad\qquad\qquad
(\textbf{likelihood vectors})
\\ \label{eq:def-Li-Wi}
L_{i}(W) 
&\coloneqq \frac{W_{i} e^{-\frac{1}{\rho} D_{i}}}{\la W_{i},e^{-\frac{1}{\rho} D_{i}} \ra},\quad \rho > 0,\qquad i \in I,
\end{align}
\end{subequations}
where $\rho$ is a user parameter to normalize the distances induced by the specific features $f_{i}$ at hand. This representation of the data is regularized by geometric smoothing \eqref{eq:def-geom-mean} to obtain the
\begin{equation}\label{eq:def-SW}
S(W) \in \mc{W},\qquad
S_{i}(W) \coloneqq \mc{G}^{w}_{i}\big(L(W)\big),\quad i \in I,
\qquad(\textbf{similarity vectors})
\end{equation}
which in turn evolves the assignment vectors $W_{i},\, i \in I$ through the
\begin{equation}\label{eq:assignment-flow}
\dot W = R_{W}\big(S(W)\big),\qquad
W(0) = \eins_{\mc{W}}.
\qquad\qquad\qquad (\textbf{assignment flow})
\end{equation}

We refer to \cite{Astrom:2017ac} and \cite{Schnorr:2019aa} for further details and a discussion of the assignment flow \eqref{eq:assignment-flow}: 
each assignment vector $W_{i}(t) \in \mc{S}$ converges to an $\veps$-neighborhood of some vertex (unit vector) $e_{j} \in \{0,1\}^{|J|},\; j \in J$ of the closure $\ol{\mc{S}}$ and in this sense uniquely assigns a corresponding label $j \in J$ to each datum $z^{i},\; i \in I$.

\subsection{Coupling the Assignment Flow and Label Evolution on Feature Manifolds}
\label{sec:Assignment-Unsupervised}
We show in this section how combining the assignment flow \eqref{eq:assignment-flow} and the schemes of Section \ref{sec:Manifold-Clustering} results in \textit{coupled flows} that \textit{simultaneously} perform 
\begin{itemize}
\item
label evolution on a feature manifold, and 
\item
spatially regularized label assignment to given data.
\end{itemize}

Coupling the assignment flow with the scheme of Section \ref{sec:soft-k-manifold} defines the \textbf{coupled flow (CFa)} in Section \ref{sec:regularized-soft-k-means-manifold}, whereas coupling the assignment flow with the scheme of Section \ref{sec:manifold-EM} defines the \textbf{coupled flow (CFb)} in Section \ref{sec:regularized-EM-manifold}. Comparing (CFa) and (CFb) in Section \ref{sec:coupled-flow-final} shows that the latter flow subsumes the former one and hence defines the \textbf{unsupervised assignment flow}.

\subsubsection{Spatially Regularized Soft-$k$-Means on Feature Manifolds}
\label{sec:regularized-soft-k-means-manifold}

Minimizing the objective function \eqref{eq:geom-soft-k-means-objective} induces the assignment probabilities
\begin{equation}\label{eq:p-eps-ij}
p^{i}_{\veps,j}(M) = \frac{\exp\big(-\frac{1}{\veps} D(z^{i},m^{j})\big)}{\sum_{l \in J} \exp\big(-\frac{1}{\veps} D(z^{i},m^{l})\big)},\quad
i \in I,\; j \in J
\end{equation}
due to \eqref{eq:djJM-manifold}. Regarding the assignment flow, the variables $W_{ij}$ play the same role, see \eqref{eq:Wij-Pr}. The assignment flow \eqref{eq:assignment-flow} for $W_{ij}$ reads
\begin{equation}
\dot W_{ij}(t) = W_{ij}(t)\Big(S_{ij}\big(W(t)\big)-\sum_{l \in J} W_{il}(t) S_{il}\big(W(t)\big)\Big),\quad i \in I,\; j \in J,
\end{equation}
where the right-hand side comprises the similarity vectors $S_{i}(W),\,i \in I$, whose $j$-th component due to \eqref{eq:def-SW}, \eqref{eq:def-geom-mean} and \eqref{eq:def-LW} is given by
\begin{subequations}\label{eq:SijW}
\begin{align}
S_{ij}(W) 
= \frac{\widetilde L_{\mc{N}_{i},j}}{\la\eins,\widetilde L_{\mc{N}_{i}} \ra},\qquad
\widetilde L_{\mc{N}_{i},j}
&= \Big(\prod_{k \in \mc{N}_{i}} L_{kj}(W;M)\Big)^{w_{ik}} 
\\
&= \Bigg(\prod_{k \in \mc{N}_{i}} \bigg(\frac{W_{kj}}{\big\la W_{k},e^{-\frac{1}{\rho} D_{k}(M)}\big\ra}\bigg)^{w_{ik}}\Bigg)
\exp\Big(\sum_{k \in \mc{N}_{i}} w_{ik} \frac{D(z^{k},m^{j})}{\rho}\Big).
\end{align}
\end{subequations}
This expression makes explicit how spatial regularization through averaging the given data (in terms of distance vectors) over local neighborhoods, is part of the vector field that drives the assignment flow. As a consequence, label assignments induced by $W(T),\; T \gg 0$, are spatially more coherent. 

Hence, we propose to replace in \eqref{eq:fp-iter-M-general} the normalized assignment variables $p^{i}_{\veps,j}(M)$ given by \eqref{eq:p-eps-ij}, where \textit{no} spatial regularization is involved, by the \textit{normalized} assignment variables $q^{j}_{\veps,i}(W)$ defined below by \eqref{eq:coupled-a}. 
The resulting \textbf{coupled flow (CFa)} that simultaneously performs label evolution and label assignment reads
\begin{equation}\label{eq:coupled-a}
\textbf{(CFa)} \quad \left\{
\begin{aligned}
\dot m^{j}(t)
&= -\alpha \sum_{i \in I} q^{j}_{\veps,i}(W) \widehat g^{-1}\big(d_{j}D(z^{i},m^{j})\big),\quad m^{j}(0)=m^{j}_{0},\quad\alpha > 0,
& j &\in J,
\\ &\qquad 
q^{j}_{\veps}(W) 
= \frac{W^{j}}{\la\eins,W^{j}\ra},\quad j \in J
\\ 
\dot W_{i}(t) &= R_{W_{i}(t)}\big(S_{i}\big(W(t)\big)\big),\quad W_{i}(0) = \eins_{\mc{S}},\quad
i \in I,
\end{aligned}
\right.
\end{equation}
with $W^{j}$ due to \eqref{eq:def-assignment-matrix} and user parameter $\alpha$ that enables the adjust the time scale of the label flow induced by $\dot m^{j}(t),\; j \in J$ relative to the assignment flow induced by $\dot W_{i}(t),\; i \in I$.

\subsubsection{Spatially Regularized EM-Iteration on Feature Manifolds}
\label{sec:regularized-EM-manifold}

The scheme of Section \ref{sec:manifold-EM} and the update formulas \eqref{eq:EM-manifold} suggest an alternative coupling of label evolution and the assignment flow. Equation \eqref{eq:def-Li-Wi} reads
\begin{equation}\label{eq:Lij-EM}
L_{ij}(W;M) = \frac{W_{ij} e^{-\frac{1}{\rho} D(z^{i},m^{j})}}{\sum_{l \in J} W_{il} e^{-\frac{1}{\rho} D(z^{i},m^{l})}},
\end{equation}
which agrees with the right-hand side of \eqref{eq:pj|i-EM}, except for the scaling parameter $\rho$ and the assignment variables $W_{ij}$ in place of the mixture coefficients $\pi_{j}$. Indeed, since there is \textit{no} interaction between different spatial locations $i \in I$ on the right-hand side of \eqref{eq:Lij-EM}, $L_{ij}(W_{i};M)$ can be interpreted as \textit{local} posterior probability of label $j$ given the observation $z^{i}$, in agreement with the left-hand side of \eqref{eq:pj|i-EM}. Likewise, 
applying the first update equation of \eqref{eq:mj-update-EM-manifold} to \eqref{eq:Lij-EM} yields 
\begin{equation}
W_{ij}^{(t+1)} = \frac{1}{|I|} \sum_{i \in I} L_{ij}(W^{(t)};M^{(t)}),\quad j \in J
\end{equation}
which does \textit{not} depend on $i \in I$. We therefore take into account spatial regularization by replacing the mixture coefficients $\pi_{j},\, j \in J$ by the variables $W_{ij},\, i \in I, j \in J$, that are governed by the assignment flow and hence \textit{do} spatially interact. The resulting \textbf{coupled flow (CFb)} reads
\begin{equation}\label{eq:coupled-b}
\textbf{(CFb)}\quad\left\{
\begin{aligned}
\dot m^{j}(t)
&= -\alpha \sum_{i \in I} \nu_{j|i}\big(W(t), M(t)\big) \widehat g^{-1}\big(d_{j}D(z^{i},m^{j}(t))\big),\quad m^{j}(0)=m^{j}_{0},\;\alpha > 0,\;
j \in J,
\\ 
&\qquad
\nu_{j|i}(W,M) = \frac{L_{ij}(W;M)}{\sum_{k \in I}L_{kj}(W;M)},\quad
L_{ij}(W;M) = \frac{W_{ij} e^{-\frac{1}{\rho} D(z^{i},m^{j})}}{\sum_{l \in J} W_{il} e^{-\frac{1}{\rho} D(z^{i},m^{l})}},
\\ 
\dot W_{i}(t) &= R_{W_{i}(t)}\big(S_{i}\big(W(t)\big)\big),\quad W_{i}(0) = \eins_{\mc{S}},\quad
i \in I,
\end{aligned}
\right.
\end{equation}
where $W(t)$ depends on $M$ through \eqref{eq:SijW}.

\subsubsection{Unsupervised Assignment Flow}
\label{sec:coupled-flow-final}

We examine the relation between the coupled flows (CFa) \eqref{eq:coupled-a} and (CFb) \eqref{eq:coupled-b}. Comparing $L_{ij}(W;M)$ given by \eqref{eq:coupled-b} with $q^{j}_{\veps,i}(W)$ given by \eqref{eq:coupled-a} shows due to $\sum_{l \in J} W_{il}=1,\; i \in I$ and
\begin{subequations}\label{eq:L_{ij}-to-qij}
\begin{align}
W_{ij} &= \lim_{\rho \to \infty} L_{ij}(W;M)
\intertext{that}
q^{j}_{\veps,i}(W) 
&=  \lim_{\rho \to \infty} \nu_{j|i}(W,M).
\end{align}
\end{subequations}
We conclude that \textbf{(CFa) is a special case of (CFb)}. Since the scaling parameter $\rho$ plays a unique role in \eqref{eq:def-LW}, however, we propose to parameterize $L_{ij}(W;M)$ of \eqref{eq:coupled-b} in the \textit{same} way, but with another \textit{independent} parameter $\sigma > 0$ replacing $\rho$, in order to `interpolate' smoothly between the coupled flows (CFa) and (CFb) in the sense of \eqref{eq:L_{ij}-to-qij}.

As a result, the final form of our approach, called \textbf{unsupervised assignment flow (UAF)}, reads
\begin{equation}\label{eq:UAF}
\textbf{(UAF)}\quad\left\{
\begin{aligned}
\dot m^{j}(t)
&= -\alpha \sum_{i \in I} \nu_{j|i}\big(W(t),M(t)\big) \widehat g^{-1}\big(d_{j}D(z^{i},m^{j}(t))\big),\quad m^{j}(0)=m^{j}_{0},\quad\alpha > 0,\;
j \in J,
\\ 
&\qquad
\nu_{j|i}(W,M) = \frac{L^{\sigma}_{ij}(W;M)}{\sum_{k \in I}L^{\sigma}_{kj}(W;M)},\quad
L^{\sigma}_{ij}(W;M) = \frac{W_{ij} e^{-\frac{1}{\sigma} D(z^{i},m^{j})}}{\sum_{l \in J} W_{il} e^{-\frac{1}{\sigma} D(z^{i},m^{l})}},\quad
\sigma > 0,
\\ 
\dot W_{i}(t) &= R_{W_{i}(t)}\big(S_{i}\big(W(t)\big)\big),\quad W_{i}(0) = \eins_{\mc{S}},\quad
i \in I,
\end{aligned}
\right.
\end{equation}
with user parameters $\alpha > 0$ controlling the relative speed of label vs.~assignment evolution, and parameter $\sigma > 0$ as just discussed.

As already mentioned in Section \ref{sec:Metric-Clustering}, the greedy $k$-center clustering provides an overcomplete set of labels in a preprocessing step which is used as an initial condition $\{m^{j}_{0}\}_{j \in J}$ for the prototype component of the unsupervised assignment flow (\ref{eq:UAF}).

\subsection{Geometric Numerical Integration}\label{sec:geoNumInt}
In this subsection, we detail the iterative scheme that is used in Section \ref{sec:Feature-Manifolds} for numerically integrating the unsupervised assignment flow (\ref{eq:UAF}). We rewrite these equations more compactly in the form
\begin{subequations}
\begin{align}
\dot{W}_i(t) &= R_{W_i(t)} F_{i} \big( W(t), M(t) \big), \quad W_i(0) = \eins_{\mathcal{S}}, \quad i \in I, \label{eq:flow-W}\\
\dot{m}^j(t) &= G_{j} \big( W(t), M(t) \big), \quad m^j(0) = m_0^j, \quad j \in J, \label{eq:flow-M}
\end{align}
\end{subequations}
where the dependency of $F_i (W, M) = S_i(W)$ on $M$ is implicitly given via the distance vectors (\ref{eq:def-Di}) and the dependency of the similarity vectors on these distance vectors -- cf. (\ref{eq:def-LW}) and (\ref{eq:def-SW}).

In order to uniformly evaluate our approach for various feature manifolds
$\mathcal{M}$, we simply use the Riemannian \textit{explicit} Euler scheme for integrating the prototype evolution flow (\ref{eq:flow-M}), i.\,e., 
\begin{equation} \label{eq:explicit-geo-euler}
(m^j)^{(t+1)} = \exp_{ \mc{M}, (m^j)^{(t)}} \Big( h G_j \big( W^{(t)}, M^{(t)} \big) \Big), \quad j \in J,
\end{equation}
with step size $h > 0$, and $\exp_{ \mc{M}, m^j }$ is defined by \eqref{eq:exp-Riemannian} for  
the Riemannian manifold $\mathcal{M}$. 
In order to numerically integrate the assignment flow (\ref{eq:flow-W}), we adapt the geometric \textit{implicit} Euler scheme from \cite{Zeilmann:2018aa}. It amounts to solving the fixed point equation
\begin{equation}
V^{(t+1)} = h \Pi_{ \mathcal{T}_0 } F \Big( \exp_{ W^{(t)} } \big( V^{(t+1)} \big), M^{(t+1)} \Big),
\end{equation}
by an iterative inner loop, where $\Pi_{ \mathcal{T}_0 }$ denotes the orthogonal projection onto the tangent space \eqref{eq:def-T0-product}, followed by updating
\begin{equation}
W^{(t+1)} = \exp_{ W^{(t)} } \big( V^{(t+1)} \big).
\end{equation}
Here, $\exp_{W}$ denotes the map given by \eqref{eq:def-exp-1} and \eqref{eq:assignment-flow-maps}.

\section{Case Studies: Label Learning on Feature Manifolds}
\label{sec:Feature-Manifolds}

In the preceding section, we derived the unsupervised assignment flow \eqref{eq:UAF} for a general feature manifold $\mc{M}$ together with a geometric numerical integration scheme. In the following three subsections, we illustrate the approach by working out  details of three concrete feature manifolds. These scenarios will be evaluated numerically in the experiments section \ref{sec:Experiments}.

\subsection{$\LG{SO}(3)$-Valued Image Data: Orthogonal Frames in $\R^3$}
\label{sec:SO3}
In this subsection, we study clustering on the Lie group $\mathrm{SO}(n)$ of $n \times n$ rotation matrices. This is a smooth Riemannian manifold whose tangent space at $R \in \mathrm{SO}(n)$ is given by
\begin{equation}
T_R \mathrm{SO}(n) = \{ R\Omega \colon \Omega \in \mathfrak{so}(n) \},
\end{equation}
where $\mathfrak{so}(n) = \{ \Omega \in \R^{n \times n} \colon \Omega^\top = - \Omega \}$ denotes the Lie algebra of $\mathrm{SO}(n)$, and with the Riemannian metric given by the Frobenius inner product $g_R(A_1, A_2) = \tr(A_1^\top A_2)$. Based on the matrix exponential $\expm$ and logarithm $\logm$ \cite{Higham:2008aa}, the corresponding exponential and logarithmic maps read
\begin{equation}
\exp_{\mathrm{SO}(n),R}(R\Omega) = R \expm(\Omega), \quad \log_{\mathrm{SO}(n),R_1}(R_2) = R_1 \logm(R_1^\top R_2),
\end{equation}
and the Riemannian distance is given by
\begin{equation}
d_{\mathrm{SO}(n)} (R_1, R_2) = \| \logm(R_1^\top R_2) \|_F.
\end{equation}
In the specific case \mbox{$n=3$}, well known formulas in closed form are available \cite{Higham:2008aa}. By Rodrigues' formula, the matrix exponential of $A \in \mathfrak{so}(3)$ is given by
\begin{equation}
\expm(A) = I + \sinc(a) A + \tfrac 12 \sinc^2(\tfrac a2) A^2,\qquad a = \sqrt{\tfrac 12 \tr(A^\top A)},
\end{equation}
with the $\sinc$-function
\begin{equation}
\sinc(x) = \begin{cases} \frac{\sin(x)}{x},& x \neq 0 \\ 1, &x=0. \end{cases}
\end{equation}
The matrix logarithm of $R \in \mathrm{SO}(3)$ with $\tr(R) = 1 + 2 \cos(\theta)$, $|\theta| < \pi$ is given by
\begin{equation}
\logm(R) = \frac{1}{2 \sinc(\theta)} (R - R^\top).
\end{equation}
Moreover, the Riemannian distance can be evaluated without computing the matrix logarithm or an eigenvalue decomposition as
\begin{equation}
d_{\mathrm{SO}(3)} (R_1, R_2) = \sqrt{2} \arccos\left( \frac{\tr(R_1^\top R_2) - 1}{2} \right), \quad R_1, R_2 \in \mathrm{SO}(3).
\end{equation}
Regarding the clustering of data $\{ R_i \}_{i \in I} \subset \mathrm{SO}(n)$, we use the canonical divergence function $D(R_1, R_2) = \tfrac 12 d_{\mathrm{SO}(n)} (R_1,R_2)^2$. As a result, the  flow of the \textbf{(UAF)} \eqref{eq:UAF} for the prototypes $S_j \in \mathrm{SO}(n)$ takes the form
\begin{equation}
\dot{S}_j(t) = \alpha \sum_{i \in I} \nu_{j|i}\big( W(t), S(t) \big) \Log_{S_j(t)}(R_i) = \alpha \sum_{i \in I} \nu_{j|i}\big( W(t), S(t) \big) S_j(t) \logm(S_j(t)^\top R_i), \quad j \in J.
\end{equation}
Discretizing this flow due to \eqref{eq:explicit-geo-euler} yields the multiplicative update scheme
\begin{equation}
S_j^{(t+1)} = S_j^{(t)} \expm \left( \alpha h \sum_{i \in I} \nu_{j|i}^{(t)} \logm\Big( \big( S_j^{(t)} \big)^\top R_i \Big) \right), \quad j \in J.
\end{equation}

\subsection{Orientation Vector Fields}
\label{sec:S1}
We consider the task of clustering orientation vector fields in the two-dimensional space. We regard these vector fields as maps from the image domain into the angle space $\bigslant{\R}{\pi\Z}$, i.\,e., we identify the line $\{ \lambda (\cos\theta, \sin\theta)^\top \colon \lambda \in \R \} \subset \R^2$ with the angle $\theta \in [0,\pi)$. Let $q\colon \R \rightarrow \bigslant{\R}{\pi\Z}$ be the quotient map $\theta \mapsto \theta \mod \pi$. Rather than operating directly on the quotient manifold $\bigslant{\R}{\pi\Z}$, we work with representatives of its elements in $\R$. In particular, a flow on the quotient manifold will be given by $q(\vartheta(t))$, where $\vartheta(t)$ is a flow in $\R$. For any two representatives $\theta_1, \theta_2 \in \R$, the induced distance is given by
\begin{equation}\label{eq:distance-orientation}
d(\theta_1,\theta_2) = d_{\mc{M}}(q(\theta_1), q(\theta_2)) = \min_{\varphi \in \pi \Z} | \theta_1 - \theta_2 + \varphi | \in [0, \tfrac{\pi}{2}],
\end{equation}
and we have $d(\theta_1,\theta_2) = 0$ if and only if $q(\theta_1) = q(\theta_2)$. For the unsupervised assignment flow (\ref{eq:UAF}), we choose the canonical divergence function $D(x,y) = \tfrac 12 \big( d_{\mc{M}}(x,y) \big)^2$ on $\mc{M}$. By \eqref{eq:distance-orientation}, this corresponds to the dissimilarity function $D(\theta_1, \theta_2) = \tfrac 12 \big( d(\theta_1, \theta_2) \big)^2$ for representatives $\theta_1, \theta_2 \in \R$. This dissimilarity function is differentiable if $q(\theta_1) \neq q(\theta_2+\tfrac{\pi}{2})$, i.\,e., if the minimizer $\varphi^* \in \argmin_{\varphi \in \pi\Z} |\theta_1 - \theta_2 + \varphi |$ is unique. In this case, we have $\frac{\partial}{\partial \theta_2} D(\theta_1, \theta_2) = \theta_2 - \theta_1 - \varphi^*$. Now, denoting $\{ \theta_i \}_{i \in I} \subset \R$ the representatives of given orientations at pixels $i \in I$ and denoting by $\{ \vartheta_j \}_{j \in J} \subset \R$ the representatives of the prototype orientations (labels), the label evolution of (\ref{eq:UAF}) takes the form
\begin{equation}\label{eq:flow-S1}
\dot{\vartheta}_j(t) = \alpha \cdot \left( \sum_{i \in I} \nu_{j|i}\big( W(t), \vartheta(t) \big) \big( \theta_i - \varphi_{ij}^*(t) \big) - \vartheta_j(t) \right) \ \ \text{with}\ \ \varphi_{ij}^*(t) \in \argmin_{\varphi \in \pi\Z}~\big| \theta_i - \vartheta_j(t) + \varphi \big|.
\end{equation}
Since this flow evolves in $\R^{|J|}$, it can be numerically integrated using classical integration schemes. 
As mentioned above, the corresponding prototype flow in $\mc{M} = \bigslant{\R}{\pi\Z}$ then is given by $q(\vartheta_j(t))$, $j \in J$.

\subsection{Feature Covariance Descriptors Fields}
\label{sec:Covariance-Descriptors}
We consider data given as covariance region descriptors, as introduced in \cite{tuzel2006region}. Details of the corresponding unsupervised assignment flow are worked out in Section 
\ref{eq:CD-basic}. In Section \ref{sec:Rotational-Invariance}, we generalize the representation to obtain descriptors that are invariant with respect to rotations of the image domain.

\subsubsection{Basic Approach}\label{eq:CD-basic}
We consider feature maps $f \colon I \rightarrow \R^s$ extracted from a given 2D image $u \colon I \rightarrow \R^c$ with $c$ channels by taking partial derivatives channel-wise, e.\,g. $u_{x} = \frac{\partial u}{\partial x}$ and $u_{xy} = \frac{\partial^2 u}{\partial x \partial y}$. 
A typical example used in our experiments is
\begin{equation} \label{eq:feature-map}
i \mapsto f^{i} = \left( u, u_{x}, u_{y}, u_{xx}, \sqrt{2} u_{xy}, u_{yy} \right)^{\T}\in \R^{6c}
\end{equation}
where $(x,y)^{\T}$ denote the image coordinates at pixel $i \in I$. 
The corresponding covariance descriptor $C_i$ 
with respect to a pixel neighborhood $\mc{N}(i) \subset I$ is given by
\begin{equation} \label{eq:covariance-descriptor}
C_i = \sum_{j \in \mc{N}(i)} \omega_{ij} ( f^{j} - \bar{f}^{i}) ( f^{j} - \bar{f}^{i} )^\top  + \veps \mrm{Id} \quad\text{with}\quad \bar{f}^{i} = \sum_{j \in \mc{N}(i)} \omega_{ij} f^{j},\quad 0 < \veps \ll 1,
\end{equation}
where $\omega_{i} = (\omega_{ij})_{j \in \mc{N}_{i}} \in \Delta_{|\mc{N}(i)|}$ are weights. We add the identity matrix with a very small $\veps$ to ensure that all descriptors are positive definite, which otherwise may not hold in particular cases like homogeneous ``flat'' image regions. 

Now we consider the task of clustering given covariance descriptors as points on the 
Riemannian manifold of symmetric positive definite matrices \cite{Bhatia:2006aa}
\begin{equation}
\mc{P}_s \coloneqq \big\{ X \in \R^{s \times s} \colon X = X^\top,\, \text{$X$ is positive definite} \big\}
\end{equation}
endowed with the Riemannian metric $g_X(U,V) = \tr(X^{-1} U X^{-1} V)$ on each tangent space $T_X \mc{P}_s = \big\{ U \in \R^{s \times s} \colon U^\top = U \big\}$. The Riemannian gradient of a function $F \colon \mc{P}_s \rightarrow \R$ is given by $\ggrad F(X) = X \partial F(X) X \in T_X \mc{P}_s$, where the symmetric matrix $\partial F(X)$ denotes the Euclidean gradient of $F$ at $X$, and matrices $X$ and $\partial F(X)$ are multiplied as usual.
Denoting the prototypes (labels) by $\{ \Lambda_j \}_{j \in J} \subset \mathcal{P}_s$, the label flow of (\ref{eq:UAF}) reads
\begin{equation}\label{eq:coupled-hpd}
\dot{\Lambda}_j(t) = - \alpha \sum_{i \in I} \nu_{j|i}\big( W(t), \Lambda(t) \big) \Lambda_j(t) \partial_{2} D\big(C_i, \Lambda_j(t) \big)  \Lambda_j(t), \quad j \in J,
\end{equation}
where $D(X, Y)$ is a proper divergence on $\mc{P}_s$ as discussed below, and $\partial_{2} D(X,Y)$ denotes its Euclidean gradient with respect to $Y$. In the following, we discuss possible choices of $D$.
An obvious choice is the canonical divergence induced by the Riemannian distance
\begin{equation}
D_{\mathrm{R}}(X, Y) = \frac 12 d_{\mc{P}_s}(X, Y)^2 = \frac 12 \sum_{k \in [s]} \big(\log \lambda_k(X, Y)\big)^2,
\end{equation}
which involves all generalized eigenvalues $\lambda_k(X, Y)$ of the matrix pencil $(X, Y)$. Considering that $D(C_i, \Lambda_j)$ has to be computed for each pair of datum $C_i$ and prototype $\Lambda_j$ at \emph{each point of time} when integrating the flow, the computation of the generalized eigenvalues would be very expensive computationally. As a more efficient alternative to $D_{\mathrm{R}}$, we consider the Stein divergence \cite{Sra:2013aa}
\begin{subequations}
\begin{align}
D_{\mathrm{S}}(X,Y) &= \log\det\big(\tfrac{X+Y}{2}\big) - \tfrac{1}{2} \log\det(XY), \\
\partial_{2} D_{\mathrm{S}}(X,Y) &= \frac{1}{2} \left( \big(\tfrac{X+Y}{2}\big)^{-1} - Y^{-1} \right).
\end{align}
\end{subequations}
It involves the determinant and the inverse of a positive definite matrix, which both can be efficiently computed using the Cholesky decomposition.
Moreover, it is shown in \cite{Sra:2013aa} that $D_{\mathrm{S}}$ is a squared distance.
Based on the choice $D=D_{S}$,  equation~(\ref{eq:coupled-hpd}) takes the form
\begin{equation} \label{eq:prototype-flowS}
\dot{\Lambda}_{j}(t) = \frac{\alpha}{2} \Big( \Lambda_j(t) - \Lambda_j(t) Q_j(t) \Lambda_j(t) \Big) \quad\text{with}\quad Q_j(t) = \sum_{i \in I} \nu_{j|i}\big( W(t), \Lambda(t) \big) \left( \frac{C_i + \Lambda_j(t)}{2} \right)^{-1}.
\end{equation}
Taking the exponential map $\exp_{\mc{P}_s, X}(U) = X^{\frac 12} \expm\big( X^{-\frac 12} U X^{-\frac 12} \big) X^{\frac 12}$ into account, with $\operatorname{expm}$ denoting the matrix exponential, and discretizing the flow with the Riemannian explicit Euler scheme (\ref{eq:explicit-geo-euler}), gives the prototype update for the Stein divergence
\begin{equation} \label{eq:prototype-flowS_discrete}
\Lambda_j^{(t+1)} = \tilde{\Lambda}_j \operatorname{expm} \left( \frac{\alpha h}{2} \Big(I - \tilde{\Lambda}_j Q_j^{(t)} \tilde{\Lambda}_j \Big) \right) \tilde{\Lambda}_j \quad\text{with}\quad \tilde{\Lambda}_j = \Big( \Lambda_j^{(t)} \Big)^{\frac 12}.
\end{equation}

\subsubsection{Rotational Invariance}\label{sec:Rotational-Invariance}
We additionally constructed a dissimilarity function on $\mc{P}_s$ that is invariant under rotations of the image domain. In contrast to the Stein divergence, this dissimilarity function takes the special structure of covariance descriptors into account and hence depends on the underlying feature map. We consider the feature map in \eqref{eq:feature-map} as an example. 

Let $u, \tilde{u} \colon \R^2 \rightarrow \R$ denote two gray value images that are related by an Euclidean transformation
\begin{equation}
\begin{pmatrix} \tilde{x} \\ \tilde{y} \end{pmatrix} = \begin{pmatrix} \cos\theta & -\sin\theta \\ \sin\theta & \cos\theta \end{pmatrix} \begin{pmatrix} x \\ y \end{pmatrix} + \begin{pmatrix} x_0 \\ y_0 \end{pmatrix}, \quad \theta \in [0, 2\pi)
\end{equation}
of the image domain, i.\,e. $\tilde{u}(\tilde{x},\tilde{y}) = u(x,y)$. Their derivatives transform as
\begin{equation}
\begin{pmatrix} \tilde{u}_{x} \\ \tilde{u}_{y} \end{pmatrix} = R_1(\theta) \begin{pmatrix} u_{x} \\ u_{y} \end{pmatrix}, \qquad
\begin{pmatrix} \tilde{u}_{xx} \\ \sqrt{2} \tilde{u}_{xy} \\ \tilde{u}_{yy} \end{pmatrix} = R_2(\theta) \begin{pmatrix} u_{xx} \\ \sqrt{2} u_{xy} \\ u_{yy} \end{pmatrix},
\end{equation}
with rotation matrices $R_1(\theta) \in \mathrm{SO}(2)$ and $R_2(\theta) \in \mathrm{SO}(3)$ given by
\begin{equation}
R_1(\theta) = \begin{pmatrix} \cos\theta & -\sin\theta \\ \sin\theta & \cos\theta \end{pmatrix}, \qquad R_2(\theta) = \begin{pmatrix} \cos^2\theta & -\sqrt{2} \cos\theta \sin\theta & \sin^2\theta \\ \sqrt{2} \cos\theta \sin\theta & \cos^2\theta - \sin^2\theta & -\sqrt{2} \cos\theta \sin\theta \\ \sin^2 \theta & \sqrt{2} \cos\theta \sin\theta & \cos^2\theta \end{pmatrix}.
\end{equation}
It follows that covariance descriptors of $u \colon I \rightarrow \R^c$ with the feature map (\ref{eq:feature-map}) transform as $\tilde{C} = R(\theta)  C R(\theta)^\top$, with a rotation matrix $R(\theta) \in \mathrm{SO}(s)$. Setting 
\begin{equation}\label{eq:R-subgroup}
\mc{R} \coloneqq \big\{ R(\theta) \colon \theta \in [0,2\pi) \big\}, 
\end{equation}
it turns out that $\mc{R}$ is a one-dimensional subgroup of $SO(s)$, i.\,e. $R(\theta_1 + \theta_2) = R(\theta_1) R(\theta_2)$. Eventually, we construct the rotation-invariant dissimilarity function by minimizing over the Lie group action of $\mc{R}$, i.\,e.
\begin{equation}\label{eq:stein-rotation-inv}
D_{\mathrm{S},\mc{R}}(X, Y) \coloneqq \min_{R \in \mc{R}}~D_{\mathrm{S}}\big( X, R Y R^\top \big) = \min_{R \in \mc{R}}~D_{\mathrm{S}}\big( R^\top X R, Y \big).
\end{equation}
If $\partial_{2} D_{\mathrm{S}}({R^*}^\top X R^*, Y) = ({R^*}^\top X R^* + Y)^{-1} - \tfrac 12 Y^{-1}$ is the same for all $R^* \in \argmin_{R \in \mc{R}}~D_{\mathrm{S}}\big( R^\top X R, Y \big)$, then $D_{\mathrm{S},\mc{R}}(X,Y)$ is differentiable in $Y$ and the derivative is given by $\partial_{2} D_{\mathrm{S},\mc{R}}(X,Y) = \partial_{2} D_{\mathrm{S}}({R^*}^\top X R^*, Y)$ \cite[Theorem 4.13+Remark 4.14]{bonnans2013perturbation}. This holds in particular if $R^*$ is unique for a given pair $(X,Y)$. \\
Using the divergence $D_{\mathrm{S},\mc{R}}$, equation \eqref{eq:coupled-hpd} yields the same prototype update formulas \eqref{eq:prototype-flowS} and \eqref{eq:prototype-flowS_discrete} as for the Stein divergence, except for the modification
\begin{equation}\label{eq:covariance-Q-rotation}
Q_j(t) = \sum_{i \in I} \nu_{j|i}\big( W(t), \Lambda(t) \big) \left( \frac{R_{ij}(t)^\top C_i R_{ij}(t) + \Lambda_j(t)}{2} \right)^{-1}
\end{equation}
with $R_{ij}(t) \in \argmin_{R \in \mc{R}} D_{\mathrm{S}}\big( R^\top C_i R, \Lambda_j(t) \big)$.
\begin{remark*}
We conclude this section with further comments on the invariant dissimilarity function \eqref{eq:stein-rotation-inv}.
\begin{enumerate}
\item We point out again that $\mc{R}$ due to \eqref{eq:R-subgroup} (and its existence) depends on the feature map $f$. For the specific case \eqref{eq:feature-map} considered above, a transformation of the form $\tilde{C} = R(\theta)  C  R(\theta)^\top$ exists since \textit{all} derivatives up to a given order are involved. Furthermore, $\mc{R}$ is a subgroup of $\mathrm{SO}(s)$ due to the proper normalization of the mixed derivatives (note the factor $\sqrt{2}$).
\item Evaluating \eqref{eq:stein-rotation-inv} amounts to solve a one-dimensional smooth but non-convex problem. We omit the details. 
\item The dissimilarity function $D_{\mathrm{S},\mc{R}}$ is not a divergence function as introduced in Section~\ref{sec:Divergence-Functions}, since $D_{\mathrm{S},\mc{R}}(X,Y) = 0$ does not imply $X=Y$, but only $[X]_{\mc{R}} = [Y]_{\mc{R}}$ with \mbox{$[X]_{\mc{R}} = \{ RXR^\top \colon R \in \mc{R} \}$}.
Unfortunately, this cannot be fixed by considering the quotient $\bigslant{\mc{P}_s}{\sim}$ with $X \sim Y$ if and only if $X \in [Y]_{\mc{R}}$, since $\bigslant{\mc{P}_s}{\sim}$ does not have a manifold structure (e.g., the equivalence class of the identity matrix is a singleton).
Nevertheless, we can plug in $D_{\mathrm{S},\mc{R}}$ into our approach that can be used with any differentiable dissimilarity function. The resulting prototypes are then representatives $\{ \Lambda_j \}_{j \in J} \subset \mc{P}_s$ of classes $[\Lambda_j]_{\mc{R}}$.
\item The set of pairs $(X,Y) \in \mc{P}_s \times \mc{P}_s$, for which $D_{\mathrm{S},\mc{R}}(X, Y)$ is not differentiable, is negligible \cite[Theorem 10.31]{Rockafellar:2009aa}. But even for such pairs  one can choose some optimal $R_{ij}(t)$ in \eqref{eq:covariance-Q-rotation}, such that the prototype flow remains well-defined.
\end{enumerate}
\end{remark*}

\section{Numerical Examples}
\label{sec:Experiments}

\newcommand{\showExpParameterFC}[2]{
\begin{centering}
    \begin{tabular}{ccrccc}
  \textbf{\footnotesize input}
    & & & \footnotesize \textbf{(UAF)} \boldmath$\sigma=0.001$ & \footnotesize \textbf{(CFb)} \boldmath$\sigma=\rho=0.1$  & \footnotesize \textbf{(CFa)} \boldmath$\sigma=\infty$ \\[-0.8cm]    
  \includegraphics[width=#1\textwidth]{{#2}.png}
    & & \rotatebox{90}{\parbox{#1\textwidth}{\quad\ \footnotesize \boldmath$\alpha = 0.1$ }} & 
    \includegraphics[width=#1\textwidth]{{Assignment_#2_alpha=0.100_sigma=0.001}.png} & 
    \includegraphics[width=#1\textwidth]{{Assignment_#2_alpha=0.100_sigma=0.100}.png} & 
    \includegraphics[width=#1\textwidth]{{Assignment_#2_alpha=0.100_sigma=inf}.png}\\
    & & &
    \includegraphics[width=#1\textwidth]{{Prototypes_#2_alpha=0.100_sigma=0.001}.png} & 
    \includegraphics[width=#1\textwidth]{{Prototypes_#2_alpha=0.100_sigma=0.100}.png} & 
    \includegraphics[width=#1\textwidth]{{Prototypes_#2_alpha=0.100_sigma=inf}.png}\\
  \textbf{\footnotesize \boldmath$k$-center + NN}
    & & &
    \includegraphics[width=#1\textwidth]{{AssignmentFC_#2_alpha=0.100_sigma=0.001}.png} & 
    \includegraphics[width=#1\textwidth]{{AssignmentFC_#2_alpha=0.100_sigma=0.100}.png} & 
    \includegraphics[width=#1\textwidth]{{AssignmentFC_#2_alpha=0.100_sigma=inf}.png}\\[-0.8cm] 
  \includegraphics[width=#1\textwidth]{{Assignment_#2_local}.png}
    &  \hspace{1cm} & \rotatebox{90}{\parbox{#1\textwidth}{\quad\ \footnotesize \boldmath$\alpha = 1.0$ }} & 
    \includegraphics[width=#1\textwidth]{{Assignment_#2_alpha=1.000_sigma=0.001}.png} & 
    \includegraphics[width=#1\textwidth]{{Assignment_#2_alpha=1.000_sigma=0.100}.png} & 
    \includegraphics[width=#1\textwidth]{{Assignment_#2_alpha=1.000_sigma=inf}.png}\\
    & & &
    \includegraphics[width=#1\textwidth]{{Prototypes_#2_alpha=1.000_sigma=0.001}.png} & 
    \includegraphics[width=#1\textwidth]{{Prototypes_#2_alpha=1.000_sigma=0.100}.png} & 
    \includegraphics[width=#1\textwidth]{{Prototypes_#2_alpha=1.000_sigma=inf}.png}\\
  \includegraphics[width=#1\textwidth]{{AssignmentFC_#2_local}.png}
    & & &
    \includegraphics[width=#1\textwidth]{{AssignmentFC_#2_alpha=1.000_sigma=0.001}.png} & 
    \includegraphics[width=#1\textwidth]{{AssignmentFC_#2_alpha=1.000_sigma=0.100}.png} & 
    \includegraphics[width=#1\textwidth]{{AssignmentFC_#2_alpha=1.000_sigma=inf}.png}\\[-0.8cm] 
  \textbf{\footnotesize initial prototypes}
    & & \rotatebox{90}{\parbox{#1\textwidth}{\quad\ \footnotesize \boldmath$\alpha = 5.0$ }} & 
    \includegraphics[width=#1\textwidth]{{Assignment_#2_alpha=5.000_sigma=0.001}.png} & 
    \includegraphics[width=#1\textwidth]{{Assignment_#2_alpha=5.000_sigma=0.100}.png} & 
    \includegraphics[width=#1\textwidth]{{Assignment_#2_alpha=5.000_sigma=inf}.png}\\
  \includegraphics[width=#1\textwidth]{{Prototypes_#2_local}.png}
    & & &
    \includegraphics[width=#1\textwidth]{{Prototypes_#2_alpha=5.000_sigma=0.001}.png} & 
    \includegraphics[width=#1\textwidth]{{Prototypes_#2_alpha=5.000_sigma=0.100}.png} & 
    \includegraphics[width=#1\textwidth]{{Prototypes_#2_alpha=5.000_sigma=inf}.png}\\
    & & &
    \includegraphics[width=#1\textwidth]{{AssignmentFC_#2_alpha=5.000_sigma=0.001}.png} & 
    \includegraphics[width=#1\textwidth]{{AssignmentFC_#2_alpha=5.000_sigma=0.100}.png} & 
    \includegraphics[width=#1\textwidth]{{AssignmentFC_#2_alpha=5.000_sigma=inf}.png}
    \end{tabular}
\end{centering}
}
\newcommand{\showExpParameter}[2]{
\begin{centering}
    \begin{tabular}{ccrccc}
  \textbf{\footnotesize input}
    & & & \footnotesize \textbf{(UAF)} \boldmath$\sigma=0.001$ & \footnotesize \textbf{(CFb)} \boldmath$\sigma=\rho=0.1$  & \footnotesize \textbf{(CFa)} \boldmath$\sigma=\infty$ \\[0.3cm]   
  \includegraphics[width=#1\textwidth]{{#2}.png}
    & & \rotatebox{90}{\parbox{#1\textwidth}{\centering \footnotesize \boldmath$\alpha = 0.1$}} & 
    \includegraphics[width=#1\textwidth]{{Assignment_#2_alpha=0.100_sigma=0.001}.png} & 
    \includegraphics[width=#1\textwidth]{{Assignment_#2_alpha=0.100_sigma=0.100}.png} & 
    \includegraphics[width=#1\textwidth]{{Assignment_#2_alpha=0.100_sigma=inf}.png}\\
  \textbf{\footnotesize \boldmath$k$-center + NN}
    & & &
    \includegraphics[width=#1\textwidth]{{Prototypes_#2_alpha=0.100_sigma=0.001}.png} & 
    \includegraphics[width=#1\textwidth]{{Prototypes_#2_alpha=0.100_sigma=0.100}.png} & 
    \includegraphics[width=#1\textwidth]{{Prototypes_#2_alpha=0.100_sigma=inf}.png}\\[0.2cm] 
  \includegraphics[width=#1\textwidth]{{Assignment_#2_local}.png}
    &  \hspace{1cm} & \rotatebox{90}{\parbox{#1\textwidth}{\centering \footnotesize \boldmath$\alpha = 1.0$}} & 
    \includegraphics[width=#1\textwidth]{{Assignment_#2_alpha=1.000_sigma=0.001}.png} & 
    \includegraphics[width=#1\textwidth]{{Assignment_#2_alpha=1.000_sigma=0.100}.png} & 
    \includegraphics[width=#1\textwidth]{{Assignment_#2_alpha=1.000_sigma=inf}.png}\\
    & & &
    \includegraphics[width=#1\textwidth]{{Prototypes_#2_alpha=1.000_sigma=0.001}.png} & 
    \includegraphics[width=#1\textwidth]{{Prototypes_#2_alpha=1.000_sigma=0.100}.png} & 
    \includegraphics[width=#1\textwidth]{{Prototypes_#2_alpha=1.000_sigma=inf}.png}\\[0.2cm] 
  \textbf{\footnotesize initial prototypes}
    & & \rotatebox{90}{\parbox{#1\textwidth}{\centering \footnotesize \boldmath$\alpha = 5.0$}} & 
    \includegraphics[width=#1\textwidth]{{Assignment_#2_alpha=5.000_sigma=0.001}.png} & 
    \includegraphics[width=#1\textwidth]{{Assignment_#2_alpha=5.000_sigma=0.100}.png} & 
    \includegraphics[width=#1\textwidth]{{Assignment_#2_alpha=5.000_sigma=inf}.png}\\
  \includegraphics[width=#1\textwidth]{{Prototypes_#2_local}.png}
    & & &
    \includegraphics[width=#1\textwidth]{{Prototypes_#2_alpha=5.000_sigma=0.001}.png} & 
    \includegraphics[width=#1\textwidth]{{Prototypes_#2_alpha=5.000_sigma=0.100}.png} & 
    \includegraphics[width=#1\textwidth]{{Prototypes_#2_alpha=5.000_sigma=inf}.png}
    \end{tabular}
\end{centering}
}
\newcommand{\showExpPrototypesDieOut}[2]{
\begin{centering}
    \begin{tabular}{cccc}
     \textbf{\footnotesize input} & \footnotesize \textbf{ \boldmath$k$-means}  & \footnotesize \textbf{\boldmath$k$-center}  & \footnotesize \textbf{ \textbf{(UAF)} \boldmath$|\mc N|=1$} \\[0.1cm]
       \includegraphics[width=#1\textwidth]{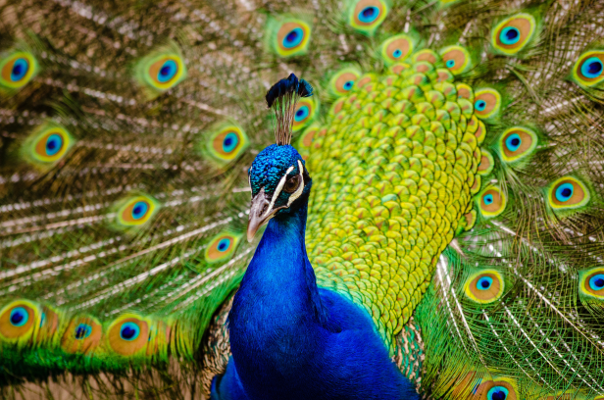} & 
       \includegraphics[width=#1\textwidth]{{Assignment_#2_kmeans}.png} & 
       \includegraphics[width=#1\textwidth]{{Assignment_#2_local}.png} & 
       \includegraphics[width=#1\textwidth]{{Assignment_#2_size_neigh=1}.png} \\
        & 
       \includegraphics[width=#1\textwidth]{{Hist_#2_kmeans}.pdf} & 
       \includegraphics[width=#1\textwidth]{{Hist_#2_local}.pdf} & 
       \includegraphics[width=#1\textwidth]{{Hist_#2_size_neigh=1}.pdf} \\[0.3cm] 
     \footnotesize \textbf{\textbf{(UAF)} \boldmath$|\mc N|=3$} & \footnotesize \textbf{ \textbf{(UAF)} \boldmath$|\mc N|=5$}  & \footnotesize \textbf{ \textbf{(UAF)} \boldmath$|\mc N|=7$}  & \footnotesize  \textbf{\textbf{(UAF)} \boldmath$|\mc N|=9$} \\[0.1cm]
       \includegraphics[width=#1\textwidth]{{Assignment_#2_size_neigh=3}.png} & 
       \includegraphics[width=#1\textwidth]{{Assignment_#2_size_neigh=5}.png} & 
       \includegraphics[width=#1\textwidth]{{Assignment_#2_size_neigh=7}.png} & 
       \includegraphics[width=#1\textwidth]{{Assignment_#2_size_neigh=9}.png} \\
       \includegraphics[width=#1\textwidth]{{Hist_#2_size_neigh=3}.pdf} & 
       \includegraphics[width=#1\textwidth]{{Hist_#2_size_neigh=5}.pdf} & 
       \includegraphics[width=#1\textwidth]{{Hist_#2_size_neigh=7}.pdf} & 
       \includegraphics[width=#1\textwidth]{{Hist_#2_size_neigh=9}.pdf}
    \end{tabular}
\end{centering}
}

In this section, we demonstrate and compare the proposed \textbf{unsupervised assignment flow (UAF)} using several synthetic and real-world images and different feature manifolds, as detailed in Section \ref{sec:Feature-Manifolds}. As described in Section \ref{sec:geoNumInt}, the geometric numerical integration of the \textbf{(UAF)} was carried out using the geometric implicit Euler scheme for the assignment component of the flow and a Riemannian explicit Euler scheme for the prototype component of the flow. For both schemes, we used the fixed step size $h=0.1$ in all experiments. Additionally we adopted in our implementation the renormalization step from \cite{Astrom:2017ac} with $\varepsilon=10^{-10}$ for the assignment component, to avoid numerical issues for assignments very close to the boundary of simplex $\Delta_{|J|}=\ol{\mc{S}}$. Uniform weights $(w_{ik})$ were used for regularizing the assignments through geometric averaging \eqref{eq:def-geom-mean}. 
The integration process terminated when the average entropy of the assignment component dropped below $10^{-3}$ which indicates almost unique assignments (probability vectors are close to unit vectors) and in turn that the weights $\nu_{j|i}(W,M)$ for the prototype evolution become stationary as well. We initialized the assignment component of the unsupervised assignment flow with the uninformative barycenter (all labels are equiprobable). The initial prototypes were determined by greedy $k$-center metric clustering as discussed in Section \ref{sec:Metric-Clustering}, in order to obtain an almost uniformly sampled dictionary from the input data. The number of labels $|J|$ was chosen large enough to start with an overcomplete dictionary.

\begin{figure}[htpb]
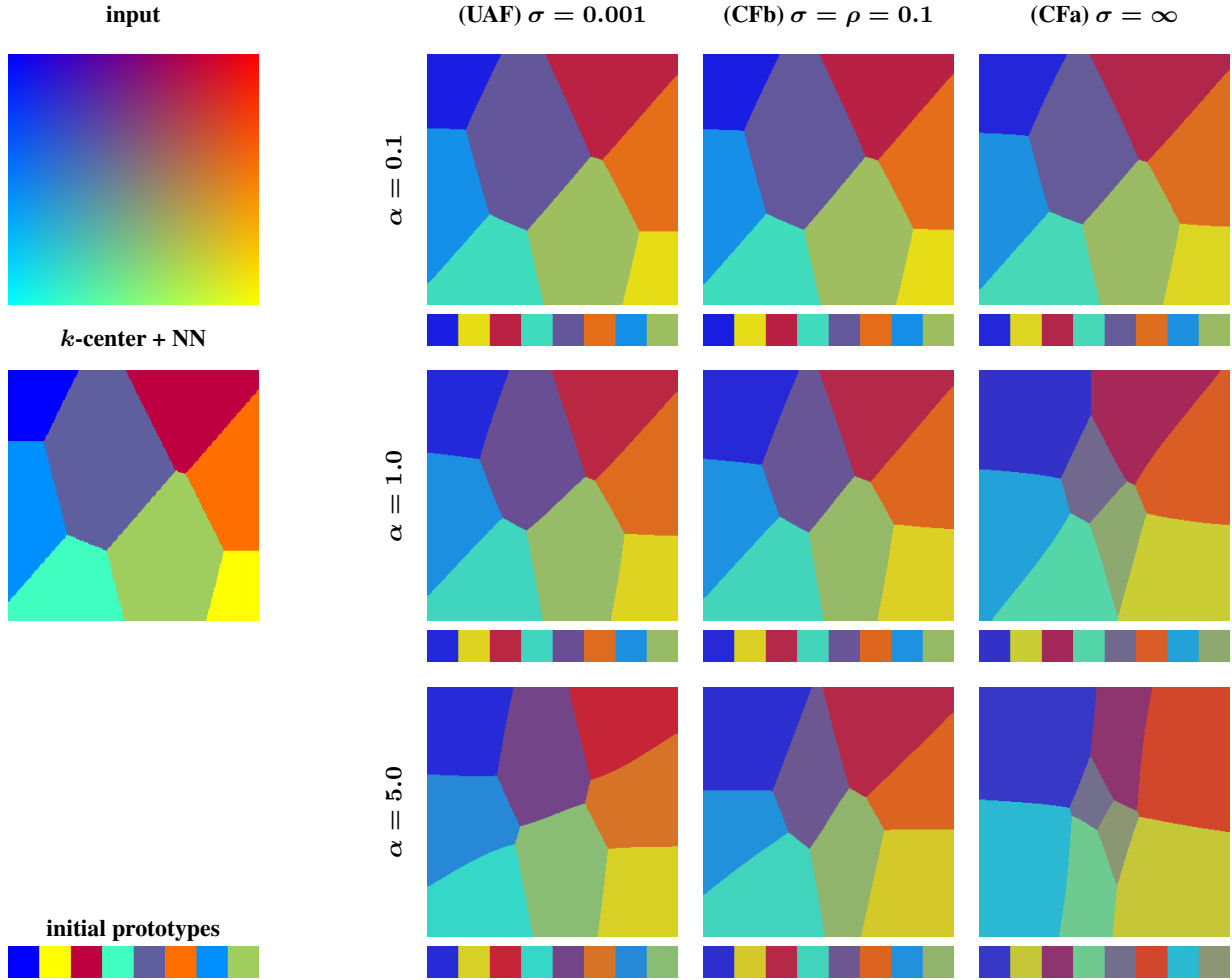

  \begin{center}
    \showExpParameter{0.2}{pnggrad8rgb}
    \caption{\textbf{Influence of the parameters \boldmath$\sigma$ and \boldmath$\alpha$.} The figure illustrates the influence of the  parameters $\sigma$ and $\alpha$ on the unsupervised assignment flow \textbf{(UAF)} in terms of the resulting labelings. From the smooth input image (left panel, top), the initial prototypes ($|J|=8$) are extracted by greedy $k$-center clustering and assigned by the nearest neighbor (NN) rule. The right panel shows the labelings returned by the \textbf{(UAF)}, for different values of $\sigma$ and $\alpha$, after termination of the coupled evolution of labels and assignments. We observe for increasing values $\sigma$ and $\alpha$ that regions are ``attracted'' toward the center of the image domain, since label colors are increasingly averaged through the spatially regularized assignments.}
    \label{fig:ExpParameter3}
  \end{center}
\end{figure}

\begin{figure}[htpb]
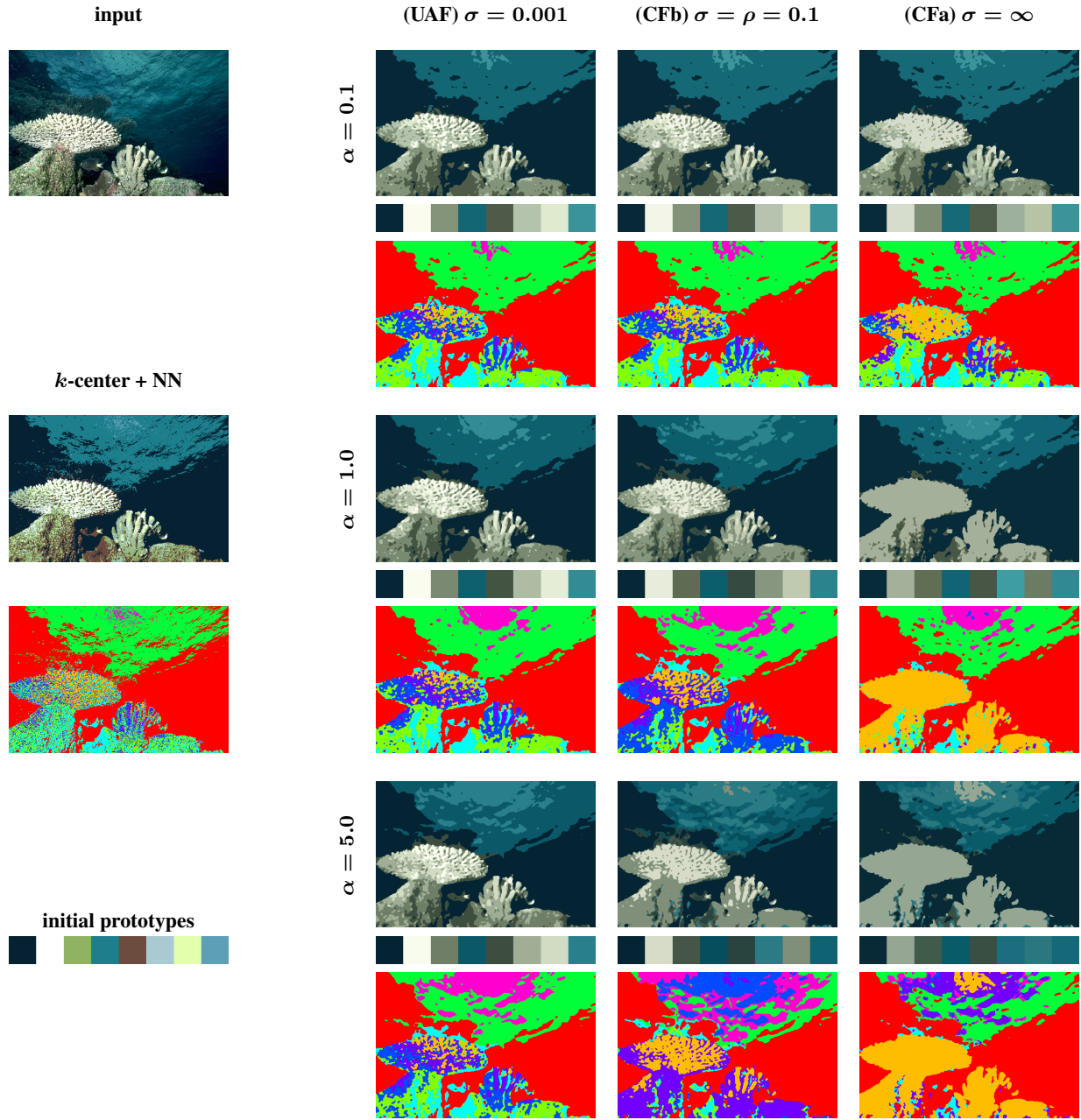

  \begin{center}
    \showExpParameterFC{0.2}{101027}
    \caption{\textbf{Influence of the parameters \boldmath$\sigma$ and \boldmath$\alpha$.} Results of the \textbf{(UAF)} are shown that reproduce for a real image the effects illustrated by Figure \ref{fig:ExpParameter3}. Each labeling is additionally shown using false colors to ease the perception of differences. We observe for increasing $\sigma$ an increasing impact of spatial regularization, whereas for increasing $\alpha$ labels adapt faster along with the size of the spatially regularized regions.}
    \label{fig:ExpParameter1}
  \end{center}
\end{figure}

\begin{figure}[htbp]
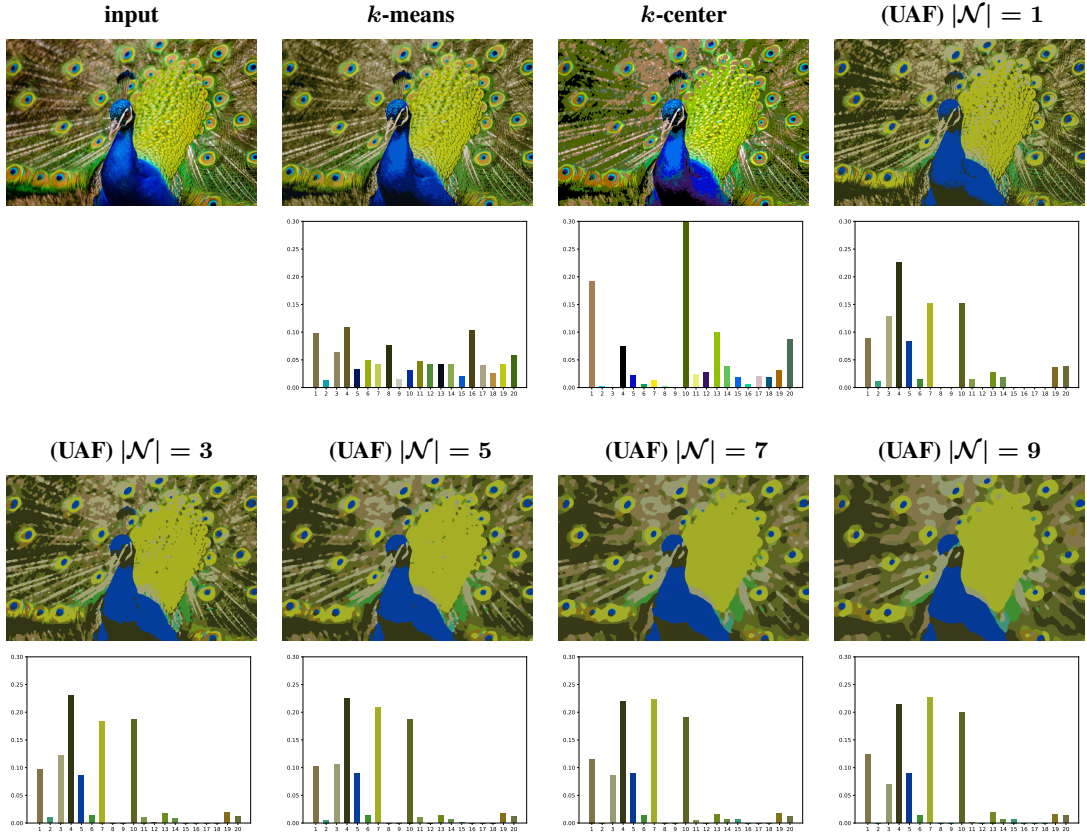

  \begin{center}
    \showExpPrototypesDieOut{0.2}{bird1}
    \caption{\textbf{Effect of spatial regularization.} We compare the proposed \textbf{(UAF)} to $k$-means clustering and $k$-center clustering, respectively, and demonstrate the effect spatial regularization, parameterized by increasing neighborhood sizes used for geometric averaging, on the resulting label statistics and label assignments. The histogram bars are colored by the corresponding labels, and their heights indicate the relative amount of assigned pixels. We observe that as the scale (neighborhood size) of spatial regularization increases, the label set quickly becomes more sparse.
    }
    \label{fig:ExpPrototypesDieOut}
  \end{center}
\end{figure}

\subsection{Parameter Influence}\label{sec:parameter-influence}
This experiment discusses the influence of the two model parameters $\sigma$ and $\alpha$ of the \textbf{(UAF)} as defined by \eqref{eq:UAF}. Parameter $\sigma$ determines the trade-off between the influence of the assignments (spatial regularization) and the influence of the distances in the feature space on the weights $\nu_{j|i}(W,M)$ which govern the label evolution. 
$\sigma=\infty$ results in the coupled flow \textbf{(CFa)} where the weights $\nu_{j|i}(W,M)$  solely depend on the assignments, whereas $\sigma=\rho$ gives coupled flow \textbf{(CFb)} which incorporates both the spatially regularized assignment and the distances in the feature space, into the dictionary update step. In general, the impact of spatial regularization on the evolution of labels evolution decreases with decreasing values of $\sigma$, and the influence of the distances in feature space on the evolution of labels is even stronger for $\sigma < \rho$. 

Parameter $\alpha$ controls the relative speed of the evolution of labels vs. assignments. If $\alpha$ is set too small, i.\,e.,  if the evolution of labels is too slow, then hardly any label evolution occurs at all during the period the assignment evolution so that the resulting assignment is effectively comparable to the \textit{supervised} assignment flow \cite{Astrom:2017ac} based on the initial set of labels. On the contrary, if $\alpha$ is set too large, labels adapt too fast to the current assignment, which may be undesirable especially if the assignment is still too close to the uninformative barycenter in the initial phase of its evolution.

In order to visualize clearly the role of $\sigma$ and $\alpha$, we consider in this section the RGB color space as feature space. 
The demonstrated effects carry over to the other non-trivial feature manifolds, of course. We used a $|\mc N|=3 \times 3$ neighborhood size for geometric spatial regularization and fixed the number of labels to $|J| = 8$.

Figure \ref{fig:ExpParameter3} illustrates the above discussion for an academic computer-generated color image with a smooth strong gradient, which was generated such that from left-to-right the red channel is increasing and the blue channel is decreasing, whereas the green channel is increasing from top to bottom. The boosted labels adaption (for larger $\alpha$), and the impact of spatial regularization is illustrated by the cell sizes of the final Voronoi diagram relative to the initial configuration. 

Figure \ref{fig:ExpParameter1} demonstrates the same effects for a real image. The partitions corresponding to the unsupervised image labelings are additionally displayed using false colors in order to highlight the differences. The interpretation of the results for different values of $\sigma$ and $\alpha$ is analogous to the effects shown by Figure \ref{fig:ExpParameter3}.

Specifically, we observe that for a small value $\sigma = 0.001$ (column \textbf{(UAF)}), which increases the influence of the distances in the feature space, the resulting labeling preserves fine scales (e.g., see left coral in Figure \ref{fig:ExpParameter1}) in comparison to the other extreme choice $\sigma = \infty$ (column \textbf{(CFa)}), where the influence of the spatial regularization through the assignments in the image domain is maximal and hence fine scales are removed from the resulting labeling. The intermediate parameter choice $\sigma = \rho = 0.1$ (column \textbf{(CFb)}) shows a good compromise between the effects caused by the two extreme values of $\sigma$. 

The influence of parameter $\alpha$ controlling the relative speed of label and assignment evolution can be seen row-wise. For small $\alpha=0.1$, the adaption of the prototypes is quite limited. For the choice $\alpha = 1.0$, we observe a good compromise between label evolution and spatial regularization through the assignment flow. Finally, a very large value $\alpha=5.0$ results in strong spatial regularization, since the labels are adapting relatively fast to the current assignments and consequently the regions assigned to labels grow faster.

\subsection{Effect of Spatial Regularization}
Figure \ref{fig:ExpPrototypesDieOut} illustrates the effect of spatial regularization performed by the \textbf{(UAF)} on the evolution of both labels and label assignments, by comparing to basic $k$-means clustering and to greedy-based $k$-center clustering (Section \ref{sec:Metric-Clustering}), respectively, where no spatial regularization is involved at all. The parameter values $\alpha=1.0$ and $\sigma = \infty$ were used.

\begin{figure}[bp]
\begin{center}
\begin{tabular}{c@{\hskip4pt}cr@{\hskip4pt}c@{\hskip4pt}c}
\footnotesize \textbf{input} & \footnotesize \textbf{ground truth} & & \footnotesize \textbf{(CFa)} & \footnotesize \textbf{(CFb)} \\
\includegraphics[width=0.18\textwidth]{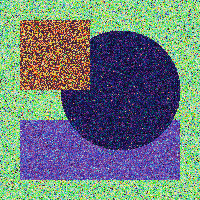} &
\includegraphics[width=0.18\textwidth]{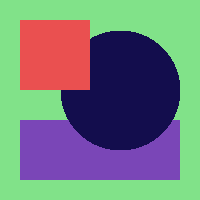} &
\qquad \rotatebox{90}{\parbox{0.18\textwidth}{\centering \footnotesize \boldmath$|\mc{N}|=1 \times 1$}} & 
\includegraphics[width=0.18\textwidth]{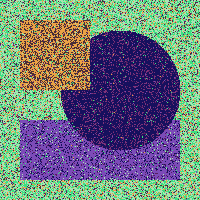} &
\includegraphics[width=0.18\textwidth]{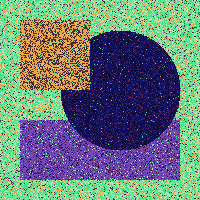} \\
\shortstack{\footnotesize \textbf{hierarchical} \\ \footnotesize \textbf{clustering}} & \footnotesize \textbf{\boldmath$k$-center + NN} & 
\quad \rotatebox{90}{\parbox{0.18\textwidth}{\centering \footnotesize \boldmath $ |\mc{N}|=3 \times 3$}} & 
\includegraphics[width=0.18\textwidth]{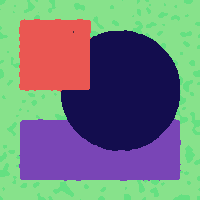} &
\includegraphics[width=0.18\textwidth]{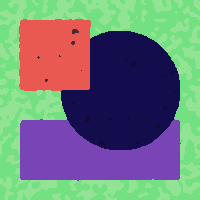} \\
\includegraphics[width=0.18\textwidth]{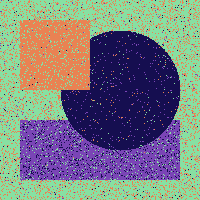} &
\includegraphics[width=0.18\textwidth]{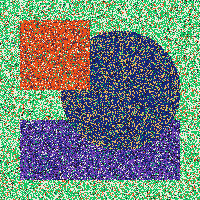} &
\qquad \rotatebox{90}{\parbox{0.18\textwidth}{\centering \footnotesize \boldmath$|\mc{N}|=5 \times 5$}} & 
\includegraphics[width=0.18\textwidth]{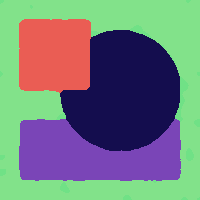} &
\includegraphics[width=0.18\textwidth]{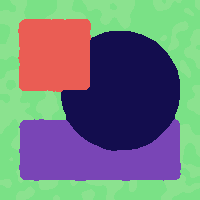}\\ 
\scriptsize (4 labels) & \scriptsize (8 labels)
\end{tabular}
\end{center}
\caption{\textbf{Unsupervised label learning for \boldmath$\mathrm{SO}(3)$-valued image data.} Rotation matrices are color coded by the scheme adopted from \cite{Kleefeld2015}. Each label (orthogonal frame, rotation matrix) is also depicted as trihedron by Figure \ref{fig:SO3_proto} using as background the false color used here. The input data were generated from ground truth as described in the text. Hierarchical clustering with generalized Ward's linkage criteria produces a noisy labeling result. The panel `$k$-center + NN' depicts the nearest neighbor assignments of the 8 labels which are selected from the input data by greedy $k$-center clustering (Section \ref{sec:Metric-Clustering}) and are used as initialization for \textbf{(UAF)}. Panels on the right depict both the labels and the assignment of these labels by the two versions \textbf{(CFa)} and \textbf{(CFb)} of the unsupervised assignment flow \textbf{(UAF)}. Spurious labels ``die out'' and, for a reasonably large neighborhood size used for spatial regularization, high-quality labelings are determined simultaneously. The resulting labels are visualized by Figure~\ref{fig:SO3_proto}.} \label{fig:SO3_seg}
\end{figure}

Comparing $k$-means with $k$-center clustering shows that $k$-means clustering selects a more uniform quantization for the feature data, whereas the greedy $k$-center clustering rather picks more extremal points in the feature space which subsequently serve as initial prototypes for \textbf{(UAF)}. The remaining panels demonstrate that spatial regularization quickly sparsifies the label set as the scale (neighborhood size) of spatial regularization increases.

\subsection{Case Studies: Label Learning on Feature Manifolds}
In this section, we demonstrate the ``plug in and play'' property of the unsupervised assignment flow \textbf{(UAF)} by applying it to the scenarios worked out in Section \ref{sec:Feature-Manifolds}. In principle, \textit{any} Riemannian feature manifold can be used provided a corresponding divergence function $D(\cdot,\cdot)$ and the exponential map admit a computationally feasible evaluation of the \textbf{(UAF)} through the numerical scheme \eqref{eq:explicit-geo-euler}.

We next consider the scenarios of Section \ref{sec:Feature-Manifolds} in turn.

\subsubsection{$\LG{SO}(3)$-Valued Image Data: Orthogonal Frames in $\R^3$}
\begin{figure}[bp]
\begin{center}
\begin{tabular}{m{5em}m{0.08\textwidth}@{\hskip4pt}m{0.08\textwidth}@{\hskip4pt}m{0.08\textwidth}@{\hskip4pt}m{0.08\textwidth}@{\hskip4pt}m{0.08\textwidth}@{\hskip4pt}m{0.08\textwidth}@{\hskip4pt}m{0.08\textwidth}@{\hskip4pt}m{0.08\textwidth}}
\footnotesize \textbf{ground truth} &
\includegraphics[width=0.08\textwidth]{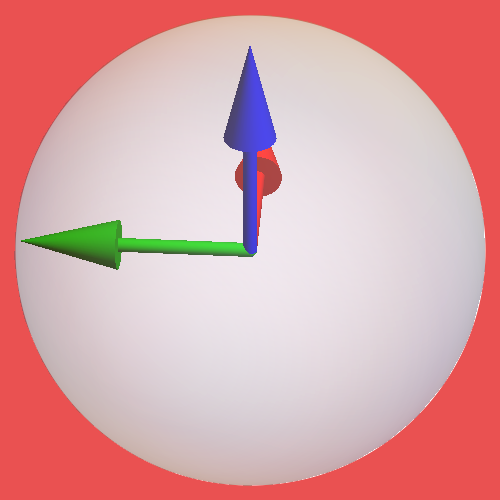} &
\includegraphics[width=0.08\textwidth]{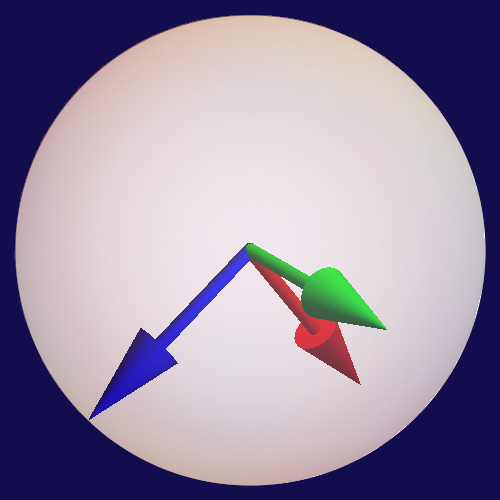} &
\includegraphics[width=0.08\textwidth]{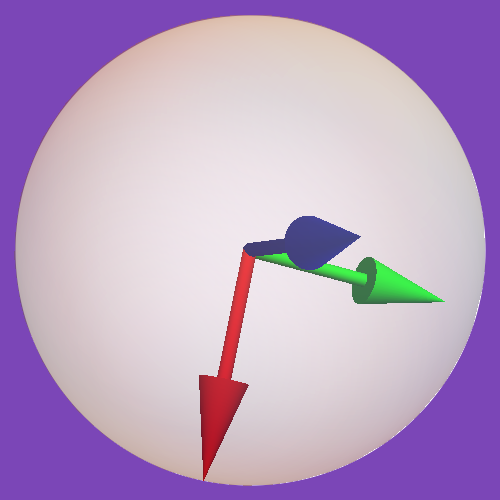} &
\includegraphics[width=0.08\textwidth]{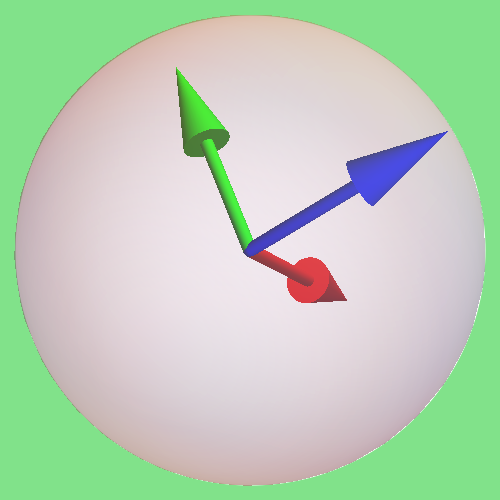} \\
\footnotesize \textbf{hierarchical clustering} &
\includegraphics[width=0.08\textwidth]{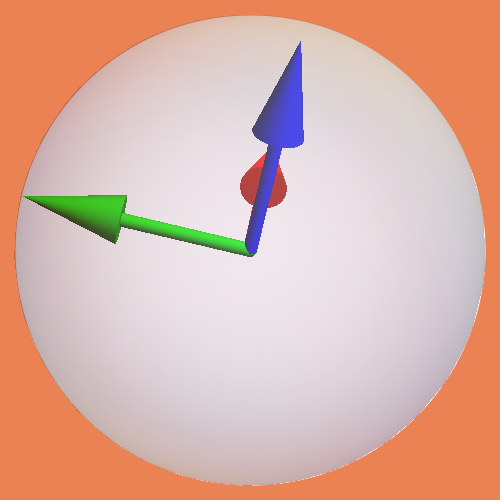} &
\includegraphics[width=0.08\textwidth]{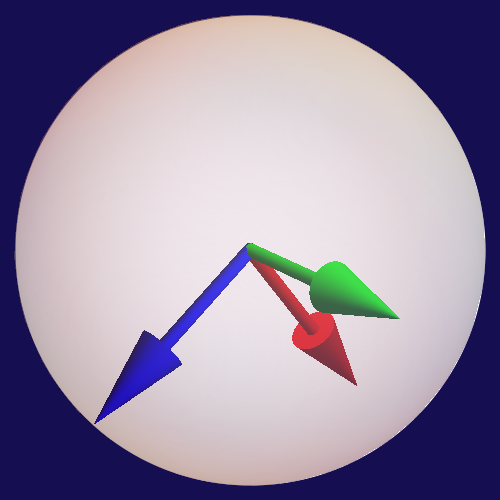} &
\includegraphics[width=0.08\textwidth]{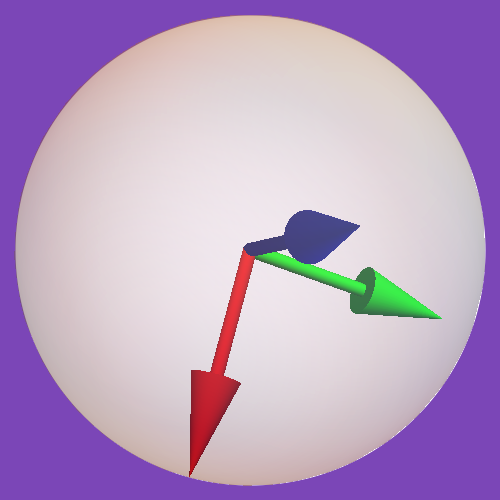} &
\includegraphics[width=0.08\textwidth]{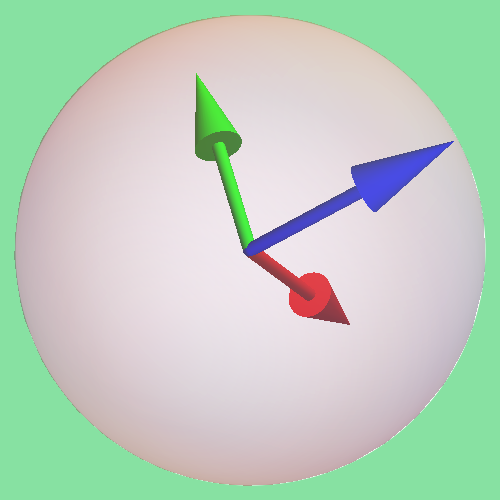} \\
\footnotesize \textbf{(CFa)} &
\includegraphics[width=0.08\textwidth]{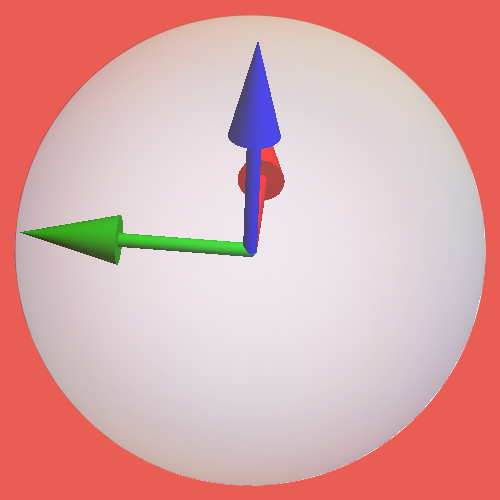} &
\includegraphics[width=0.08\textwidth]{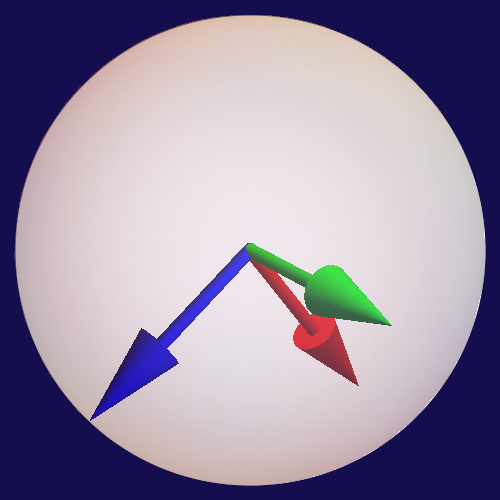} &
\includegraphics[width=0.08\textwidth]{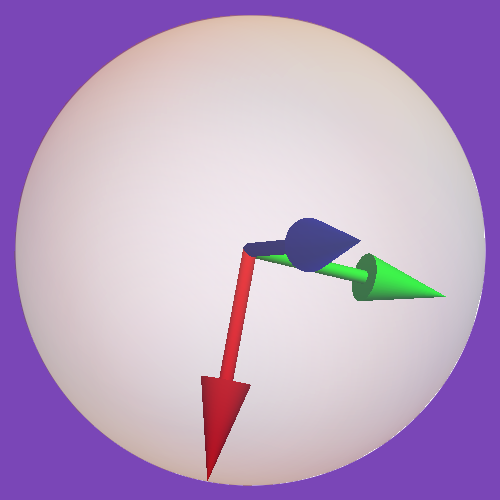} &
\includegraphics[width=0.08\textwidth]{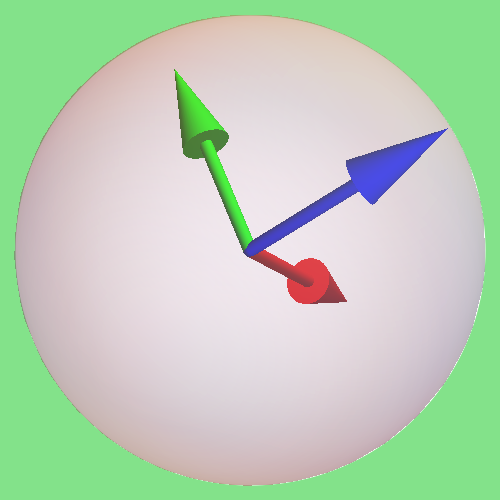} &
\includegraphics[width=0.08\textwidth]{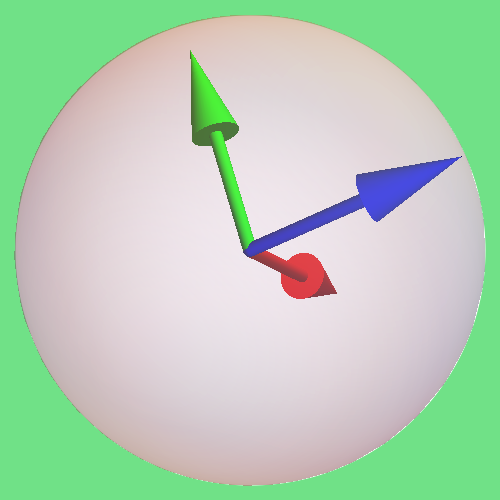} \\
\footnotesize \textbf{(CFb)} &
\includegraphics[width=0.08\textwidth]{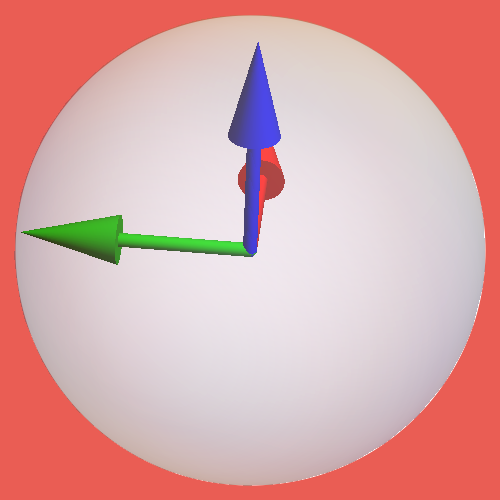} &
\includegraphics[width=0.08\textwidth]{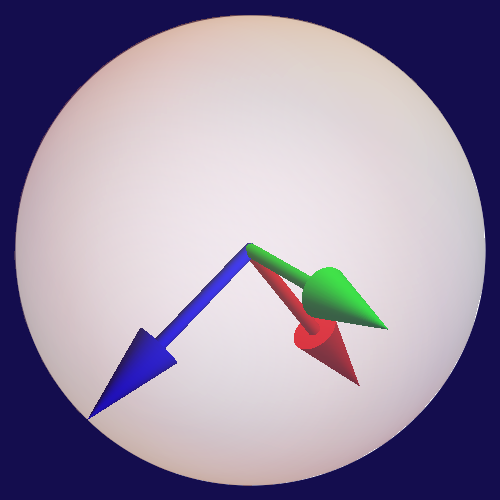} &
\includegraphics[width=0.08\textwidth]{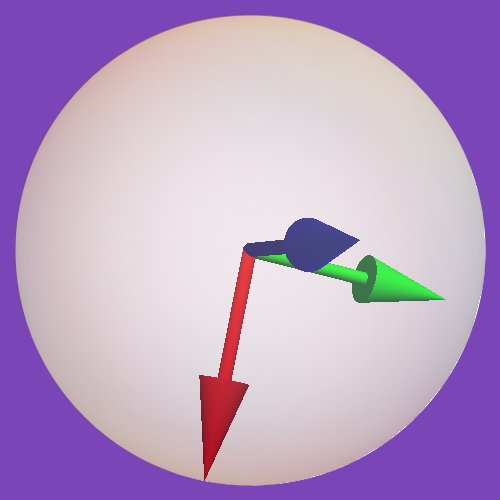} &
\includegraphics[width=0.08\textwidth]{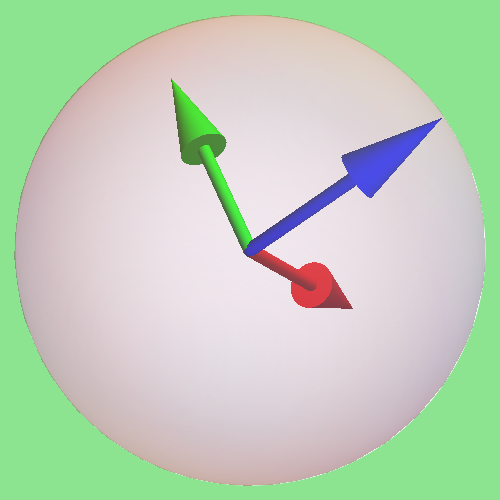} &
\includegraphics[width=0.08\textwidth]{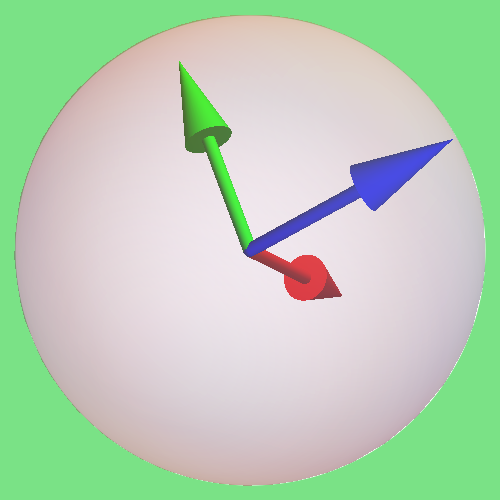} \\
& & & & & \\
\footnotesize \textbf{\boldmath$k$-center \mbox{(initialization)}} &
\includegraphics[width=0.08\textwidth]{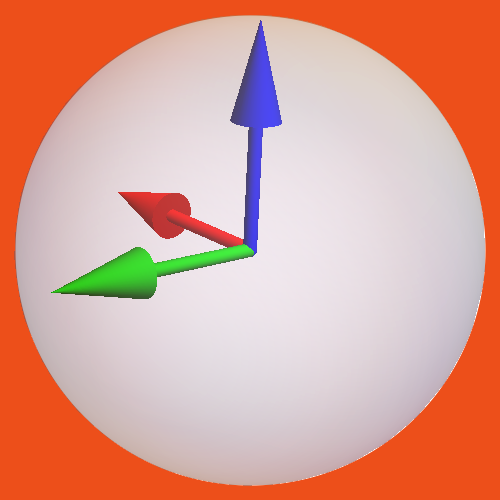} &
\includegraphics[width=0.08\textwidth]{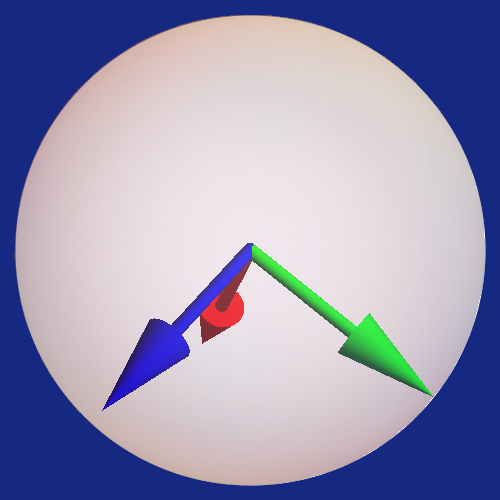} &
\includegraphics[width=0.08\textwidth]{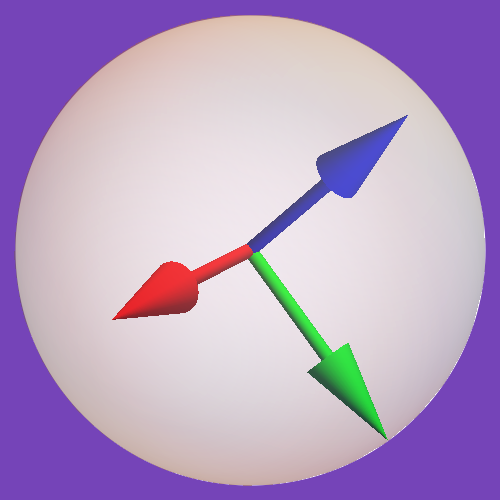} &
\includegraphics[width=0.08\textwidth]{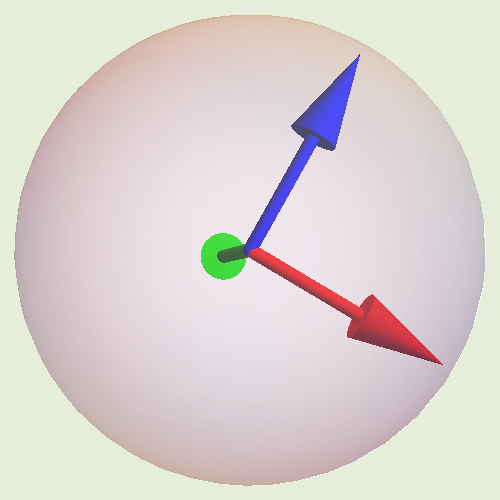} &
\includegraphics[width=0.08\textwidth]{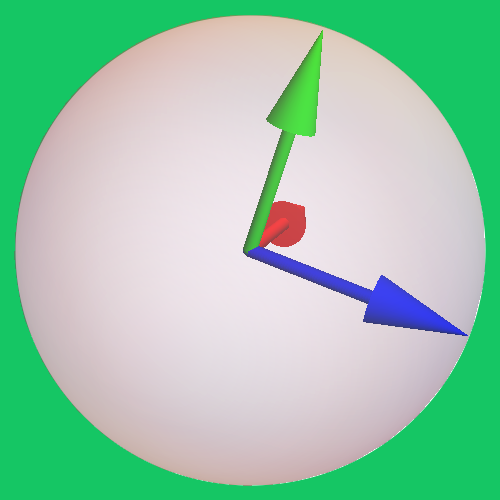} &
\includegraphics[width=0.08\textwidth]{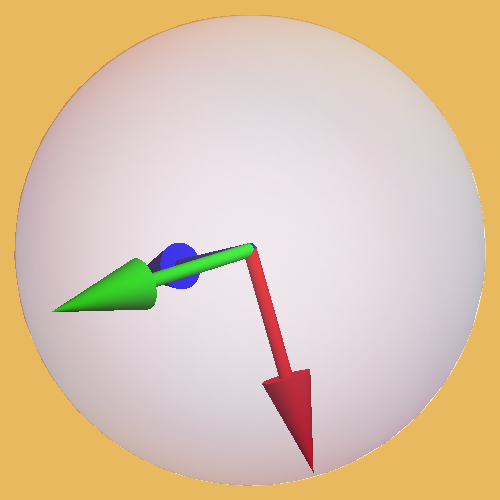} &
\includegraphics[width=0.08\textwidth]{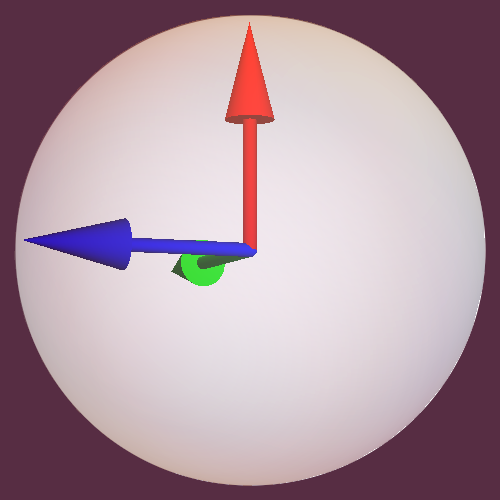} &
\includegraphics[width=0.08\textwidth]{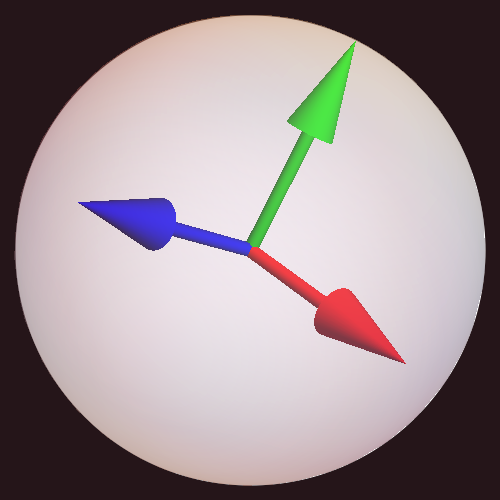}
\end{tabular}
\end{center}
\caption{\textbf{Label visualization for \boldmath$\mathrm{SO}(3)$ data.} Each label corresponding to the results depicted by Figure~\ref{fig:SO3_seg} is shown here as trihedron, using the false colors of Figure~\ref{fig:SO3_seg} as background colors here. The labels obtained by hierarchical clustering are close to the ground truth labels but also deviate significantly, as is clearly visible in the first column. In addition, the label assignments in the spatial domain cannot cope with the noise of the input data. As for \textbf{(CFa)} and \textbf{(CFb)}, three labels of the initial label set (last row) ``died out'' during the unsupervised assignment flow evolution, whereas the remaining ones converged to values quite close to ground truth. 
} \label{fig:SO3_proto}
\end{figure}

Figure \ref{fig:SO3_seg} depicts ground truth data in terms of orthogonal frames assigned to each pixel $i \in I$ and visualized with false colors. Each ground-truth label is also shown as trihedron by Figure \ref{fig:SO3_proto}. 

The input data (Figure~\ref{fig:SO3_seg}) were generated by independently sampling for each pixel $i \in I$ a vector $n_i \sim \mathcal N(0,\sqrt{0.5} I_3)$, determining a corresponding random skew-symmetric matrix $\Omega(n_i) \in \mathfrak{so}(3)$, and by replacing the ground-truth value $R_{i}$ by $R_i \expm(\Omega({n}_i))$.

We compare our method with hierarchical agglomerative clustering \cite{mullner2011modern}. As linkage criterion, we used the generalized Ward's criterion as presented in \cite{batagelj1988generalized}, i.\,e., we replaced the squared Euclidean distance in the classical Ward's method by the Riemannian distance. This linkage criterion worked best in our experiments. We chose the threshold for this method such that we get the same number of clusters as in the ground truth. The labels were determined by computing the Riemannian mean within each cluster. The noisy clustering result (Figure~\ref{fig:SO3_seg}) affects the computation of labels as can be seen in Figure~\ref{fig:SO3_proto}.

As initialization for our method, we determined by greedy-based $k$-center clustering (Section \ref{sec:Metric-Clustering}) an overcomplete set of $|J|=8$ prototypes as shown by Figure~\ref{fig:SO3_proto}. 
The corresponding nearest neighbor (NN) assignments are shown by Figure~\ref{fig:SO3_seg}. They clearly illustrate the need for spatially regularized assignments, \textit{not only} for determining a reasonably coherent partition of the image domain \textit{but also} for affecting label evolution, in order to determine proper labels enabling to find such a partition by assignment.

The labelings generated by unsupervised assignment flow \textbf{(UAF)} are shown by Figure~\ref{fig:SO3_seg}, for the parameters $\sigma=\rho=1.0$ and $\sigma=\infty$ corresponding to the specific versions \textbf{(CFa)} and \textbf{(CFb)} of the \textbf{(UAF)}, and using different neighborhood sizes $|\mc N| \in \{1\times 1, 3\times 3, 5\times 5\}$ for spatial regularization. The relative speed parameter $\alpha$ for the prototype evolution flow was set to the natural value $\alpha=1$ (cf.~Section \ref{sec:parameter-influence}). The results show that, for both flows \textbf{(CFa)} and \textbf{(CFb)}, spurious labels ``die out'' whereas the remaining labels converge to values quite close to ground truth (Figure~\ref{fig:SO3_proto}). Specifically, for the large green background region, two labels close to the ground truth label are recovered due to the initial fluctuations within  a large spatial region.

We point out that the only essential parameter value required for a reasonable result is the scale (neighborhood size) of spatial regularization.

\subsubsection{Orientation Vector Fields}
\begin{figure}[hbp]
\begin{center}
\begin{tabular}{c@{\hskip3em}r@{\hskip4pt}c@{\hskip4pt}c@{\hskip4pt}c@{\hskip4pt}cc} 
\footnotesize \textbf{original image} & & \footnotesize \textbf{input} & \footnotesize \textbf{\boldmath$k$-center + NN} & \footnotesize \textbf{(CFa)} & \footnotesize \textbf{(CFb)} \\
\includegraphics[width=0.15\textwidth]{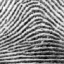} &
\rotatebox{90}{\parbox{0.15\textwidth}{\centering \footnotesize \textbf{orientation data}}} & 
\includegraphics[width=0.15\textwidth]{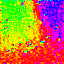} &
\includegraphics[width=0.15\textwidth]{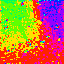} &
\includegraphics[width=0.15\textwidth]{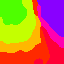} &
\includegraphics[width=0.15\textwidth]{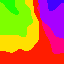} \\
& \rotatebox{90}{\parbox{0.15\textwidth}{\centering \footnotesize \textbf{overlay}}} & 
\includegraphics[width=0.15\textwidth]{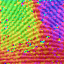} &
\includegraphics[width=0.15\textwidth]{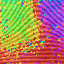} &
\includegraphics[width=0.15\textwidth]{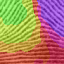} &
\includegraphics[width=0.15\textwidth]{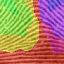} \\
& & \multicolumn{5}{l}{\hspace{-0.66em}\includegraphics[width=0.6363\textwidth]{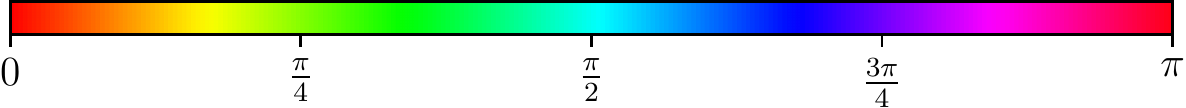}}
\end{tabular}
\caption{\textbf{Unsupervised label learning from  orientation vector fields.} Orientations are extracted from the grayscale image, using the spectral decomposition of local scatter matrices of the image gradient, and represented as elements of $\bigslant{\R}{\pi\Z} \cong S^1$ as described in the text. Using a corresponding distance function, the unsupervised assignment flow learns both proper labels, including their number, and label assignments for encoding the noisy input data.
} \label{fig:S1}
\end{center}
\end{figure}

Given a grayscale image (Figure \ref{fig:S1}) we estimated orientations of local image structure from local gradient scatter matrices. Orientations are encoded at each pixel by the angle between the horizontal axis and the smallest eigenvector. The resulting data take values in $\bigslant{\R}{\pi\Z} \cong S^1$ after identifying antipodal points. Figure \ref{fig:S1} shows the 
nearest neighbor assignments of the initial $|J|=8$ prototypes determined by greedy $k$-center clustering from the noisy input data, together with labels and label assignments of the versions \textbf{(CFa)} and \textbf{(CFb)} of the unsupervised assignment flow \textbf{(UAF)} corresponding to the parameter choices $\sigma=\rho=0.1$ and $\sigma=\infty$.
The relative speed parameter $\alpha$ for the prototype evolution was set to $\alpha=0.5$, and $|\mc{N}|=5 \times 5$ neighborhoods were used for spatial averaging. 

%

Both flows managed to position a label correctly in the neighborhood of $0 \cong \pi$ (visualized in red) and only required seven labels to properly encode the data by labeling.

\begin{figure}[htbp]
\begin{center}
\begin{tabular}{cr@{\hskip4pt}c@{\hskip4pt}c@{\hskip4pt}c}
\footnotesize \textbf{input image} & &\footnotesize \boldmath $D_{\mathrm{S}}$ & \shortstack[c]{\footnotesize \textbf{\boldmath$k$-center with \boldmath$D_{\mathrm{S},\mc{R}}$}, \\ \footnotesize \textbf{assignment with \boldmath $D_{\mathrm{S}}$}} & \footnotesize \boldmath $D_{\mathrm{S},\mc{R}}$ \\
\includegraphics[width=0.2\textwidth]{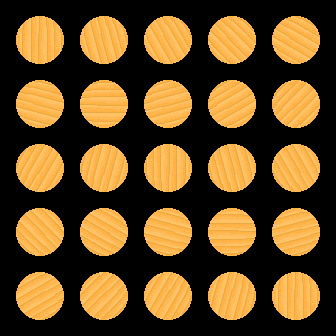} &
\qquad\rotatebox{90}{\hspace{1.6em}\footnotesize \textbf{\boldmath$k$-center + NN}} &
\includegraphics[width=0.2\textwidth]{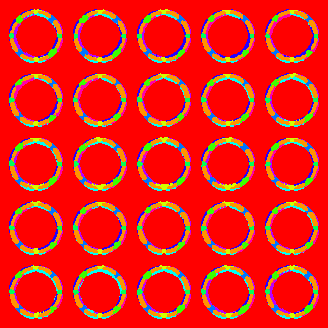} &
\includegraphics[width=0.2\textwidth]{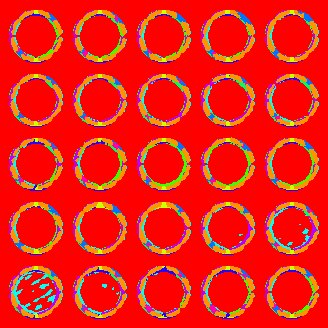} &
\includegraphics[width=0.2\textwidth]{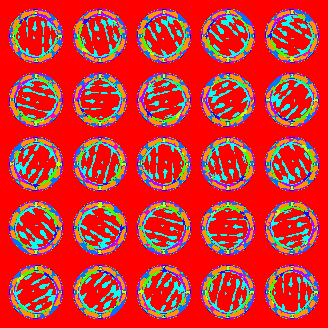} \\
&
\rotatebox{90}{\hspace{3.3em}\footnotesize \textbf{(CFa)}} &
\includegraphics[width=0.2\textwidth]{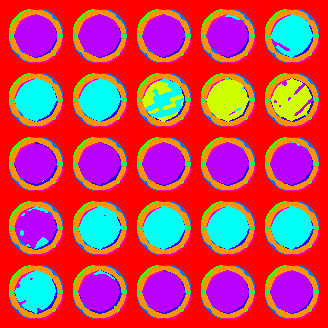} &
\includegraphics[width=0.2\textwidth]{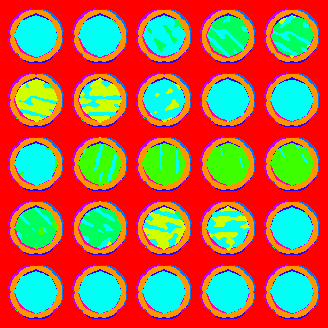} &
\includegraphics[width=0.2\textwidth]{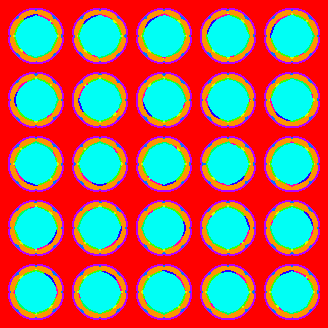}
\end{tabular}
\end{center}
\caption{\textbf{Unsupervised learning of rotationally invariant labels from covariance descriptors.} The input data are covariance descriptors \eqref{eq:covariance-descriptor} extracted from the input image which comprises a texture rotated in steps of 15 degrees. Both labels and label assignments are pixelwise visualized using false colors in the panels on the right (only color differences matter, rather than the colors themselves). The unsupervised assignment flow \textbf{(CFa)} together with the rotationally invariant Stein divergence $D_{\mathrm{S},\mc{R}}$ returns a small set of labels that encodes local image structure irrespective of its orientation.
By contrast, using the Stein divergence $D_{\mathrm{S}}$ is less effective.
}  \label{fig:cov_texture_clustering}
\end{figure}
%

\subsubsection{Feature Covariance Descriptor Fields}
We demonstrate the application of the unsupervised assignment flow to the manifold of positive definite matrices. For a given input image, we extracted the covariance descriptor using the feature map \eqref{eq:feature-map} and $|\mc{N}|=5 \times 5$ in \eqref{eq:covariance-descriptor}. We applied version \textbf{(CFa)} of the unsupervised assignment flow to a synthetic and a real world image, i.\,e. setting $\sigma = \infty$ ensuring a strong effect of spatial regularization on label evolution. Initial sets of $|J|=10$ labels were determined by metric clustering, to ensure interpretation of the results visualized by false colors. 
Due to the higher dimension of the feature space of this scenario, a larger value $\alpha=10$ of the relative speed parameter controlling the prototype evolution turned out to be useful for both test instances.

Figure \ref{fig:cov_texture_clustering} depicts a synthetic image with a texture rotated in steps of 15 degrees. $|\mc{N}|=3 \times 3$ neighborhoods were used for spatial averaging and the constant of \eqref{eq:covariance-descriptor} was set to $\varepsilon = 10^{-5}$ to ensure strict positive definiteness even in completely homogeneous regions of this computer-generated image. Initial prototypes were extracted from the input data using the greedy $k$-center clustering using the Stein divergence $D_{\mathrm{S}}$ and its rotation-invariant version $D_{\mathrm{S},\mc{R}}$, respectively. The experiments below should not only demonstrate another feature manifold that can be flexibly handled using the proposed unsupervised assignment flow, but they should also assess if numerical results display the rotational invariance of $D_{\mathrm{S},\mc{R}}$ that holds by construction mathematically (Section \ref{sec:Rotational-Invariance}).

\begin{figure}[htbp]
\begin{center}
\begin{tabular}{c@{\hskip3pt}c@{\hskip3pt}l@{\hskip3em}c@{\hskip3pt}l}
\footnotesize \boldmath $D_{\mathrm{S}}$ & \footnotesize \boldmath $D_{\mathrm{S},\mc{R}}$ & 
\multirow{2}{*}[-12.8pt]{\includegraphics[height=0.2102\textwidth]{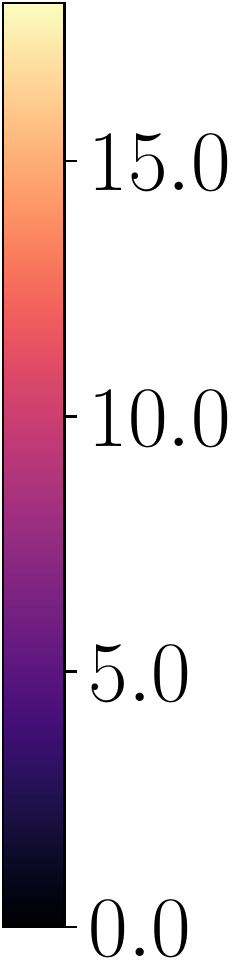}} &
\footnotesize \textbf{optimal angle} &
\multirow{2}{*}[-10.1pt]{\includegraphics[height=0.2159\textwidth]{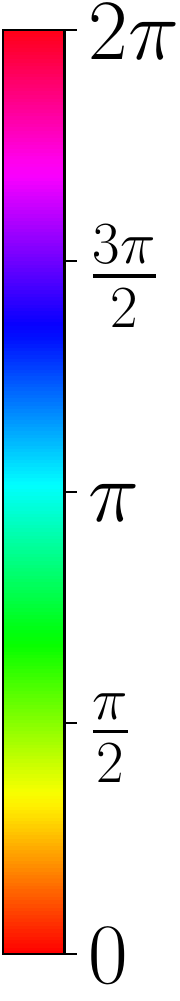}} \\
\includegraphics[width=0.2\textwidth]{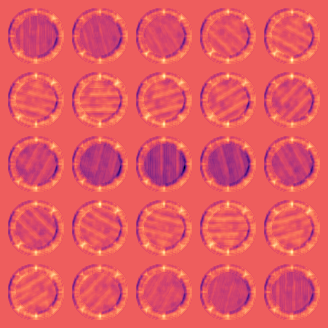} &
\includegraphics[width=0.2\textwidth]{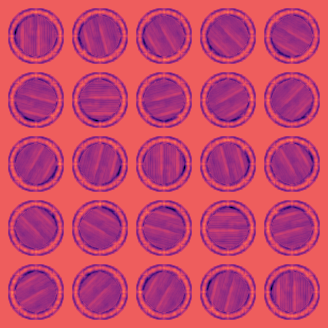} &
&
\includegraphics[width=0.2\textwidth]{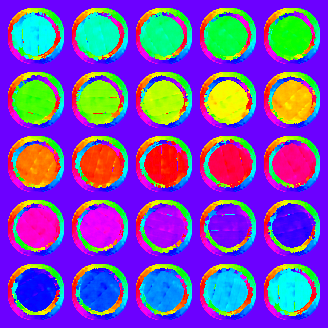} &
\end{tabular}
\end{center}
\caption{\textbf{Comparing the Stein divergence \boldmath$D_{\mathrm{S}}$ with its rotationally invariant variant \boldmath$D_{\mathrm{S},\mc{R}}$.} Using the covariance descriptors illustrated by Figure \ref{fig:cov_texture_clustering}, the panels on the left show pixelwise the distances to some fixed (arbitrary) label. Contrary to the uniform distances $D_{\mathrm{S},\mc{R}}$, the distance $D_{\mathrm{S}}$ strongly depends on the orientation of the texture. On the right-hand side, the optimal rotation angles are shown corresponding to the evaluation of $D_{\mathrm{S},\mc{R}}$. These angles accurately recover rotations of the texture.} \label{fig:cov_texture_dist}
\end{figure}
%

%
\begin{figure}[htbp]
\begin{center}
\begin{tabular}{c@{\hskip1.9em}r@{\hskip4pt}c@{\hskip4pt}c}
\footnotesize \textbf{input image} & & \footnotesize \boldmath  $D_{\mathrm{S}}$ & \footnotesize \boldmath  $D_{\mathrm{S},\mc{R}}$ \\
\includegraphics[height=0.2\textwidth]{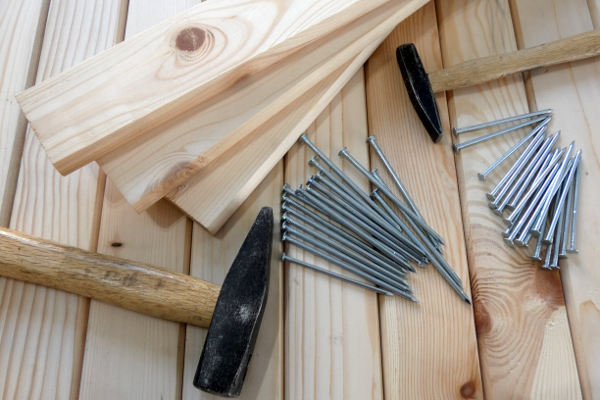} &
\rotatebox{90}{\hspace{1.6em}\footnotesize\textbf{ \boldmath$k$-center + NN}} &
\includegraphics[height=0.2\textwidth]{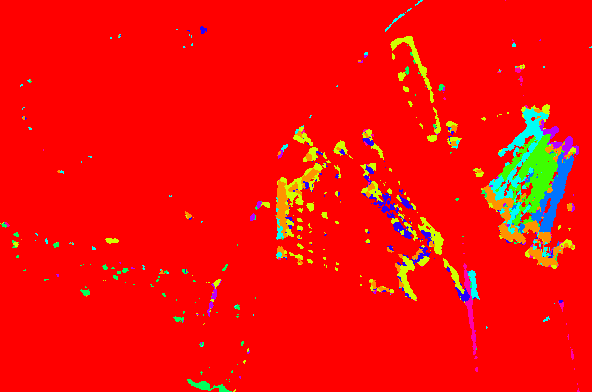} &
\includegraphics[height=0.2\textwidth]{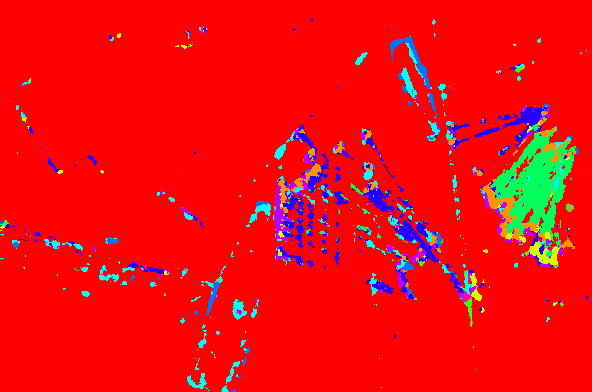} \\
&
\rotatebox{90}{\hspace{3.3em}\footnotesize \textbf{(CFa)}} &
\includegraphics[height=0.2\textwidth]{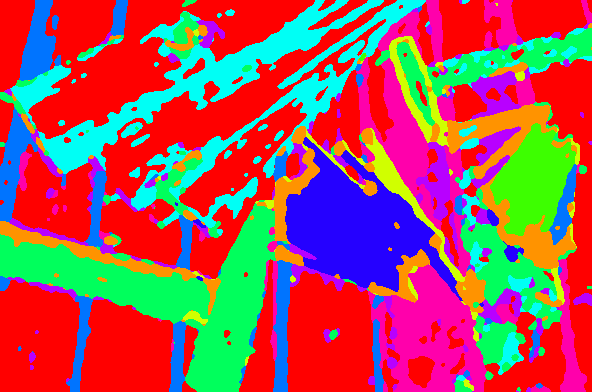} &
\includegraphics[height=0.2\textwidth]{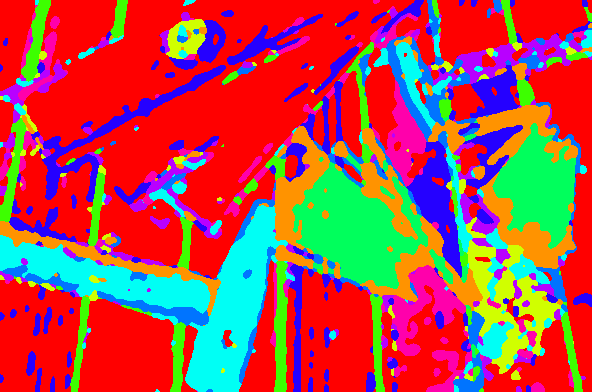}
\end{tabular}
\end{center}
\caption{\textbf{Unsupervised learning of covariance descriptor labels through label assignment.} The depicted results were obtained for the real input image on the left and are analogous to the results of the synthetic scenario depicted by Figure \ref{fig:cov_texture_clustering}. The local assignments of initial labels (top row on the right) highlight that metric clustering completely ignores the spatial structure of the input data. The results returned by the unsupervised assignment flow \textbf{(CFa)}, therefore, are impressive (bottom row): labels and label assignments jointly evolve so as to capture the spatial image structure. While the distance $D_{\mathrm{S}}$ is sensitive to orientations of texture, the distance $D_{\mathrm{S},\mc{R}}$ is not: the final labels and label assignments (bottom right) basically partition the image into wooden texture independent of the orientation of the wooden boards (encoded with red), nails and similar line structures in the background (encoded with green), the hammers (light-blue) and oriented wooden texture (blue), independent of the local orientation of these textures.
} \label{fig:cov_table_clustering}
\end{figure}
%

%
\begin{figure}[htbp]
\begin{center}
\begin{tabular}{c@{\hskip4pt}c@{\hskip4pt}l}
\footnotesize \boldmath  $D_{\mathrm{S}}$ & \footnotesize \boldmath $D_{\mathrm{S},\mc{R}}$ &
\multirow{2}{*}[-12.8pt]{\includegraphics[height=0.2102\textwidth]{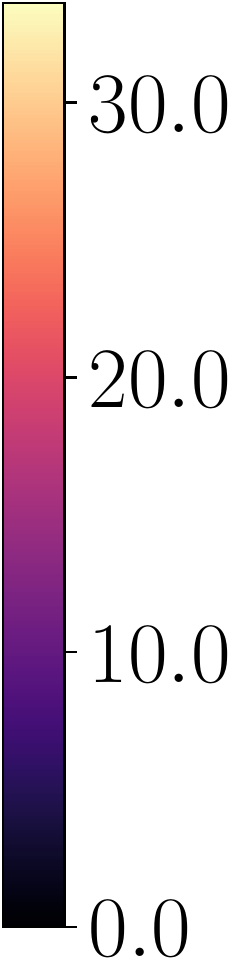}} \\
\includegraphics[height=0.2\textwidth]{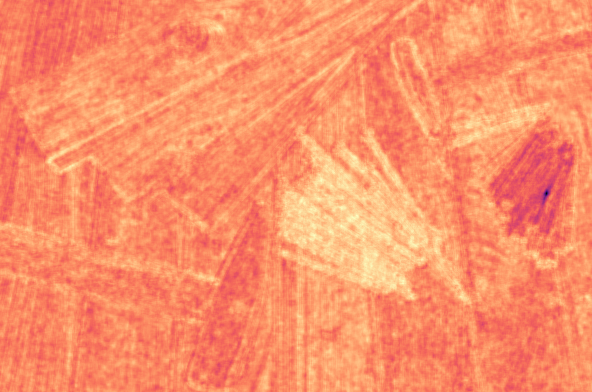} &
\includegraphics[height=0.2\textwidth]{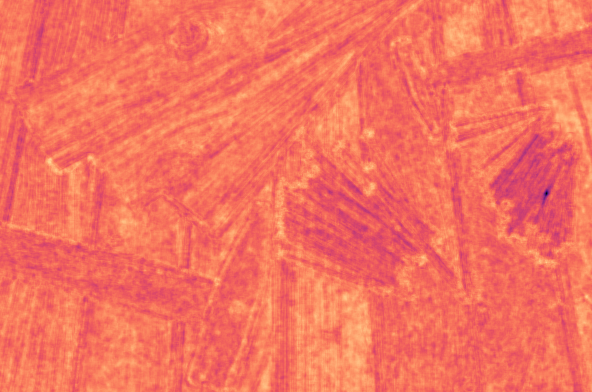} &
\end{tabular}
\\[1em]
\begin{tabular}{c@{\hskip4pt}lc@{\hskip4pt}l}
\footnotesize \textbf{optimal angle} &
\multirow{2}{*}[-10.1pt]{\includegraphics[height=0.2159\textwidth]{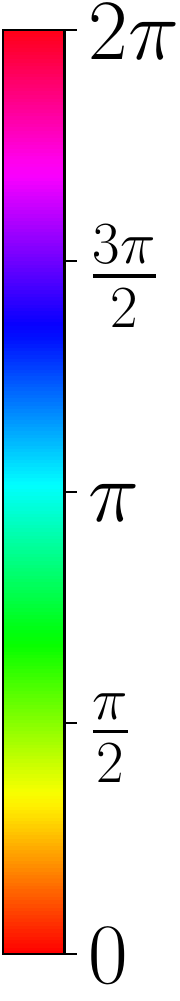}} &
\footnotesize \textbf{optimal angle modulo \boldmath$\pi$} &
\multirow{2}{*}[-10.1pt]{\includegraphics[height=0.2159\textwidth]{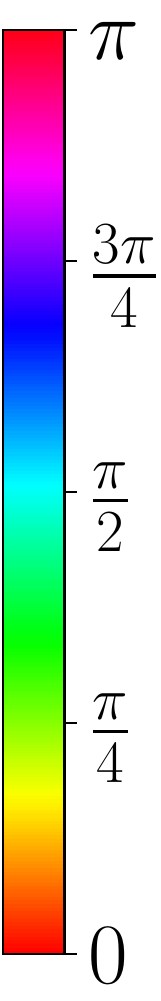}} \\
\includegraphics[height=0.2\textwidth]{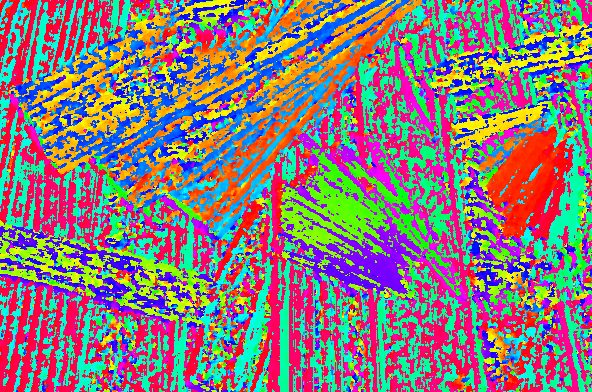} &
&
\includegraphics[height=0.2\textwidth]{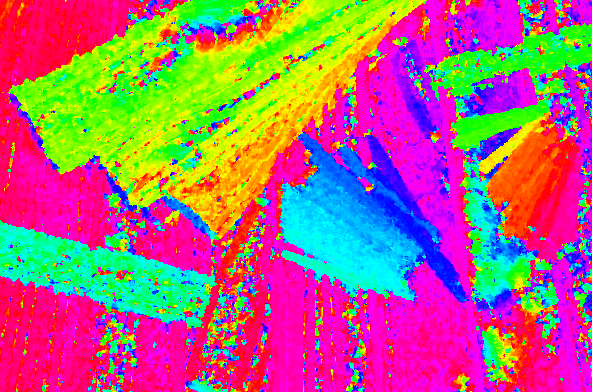} &
\end{tabular}
\end{center}
\caption{\textbf{Comparing the Stein divergence \boldmath$D_{\mathrm{S}}$ with its rotationally invariant variant \boldmath$D_{\mathrm{S},\mc{R}}$.} The depicted results correspond to the scenario of Figure \ref{fig:cov_table_clustering} and are analogous to the results shown by Figure \ref{fig:cov_texture_dist} for the synthetic scenario illustrated by Figure \ref{fig:cov_texture_clustering}. The top row shows the pixelwise distances between each covariance descriptors extracted from the image of Figure~\ref{fig:cov_table_clustering} and a fixed prototype located at the pile of nails on the right. While the distance $D_{\mathrm{S}}$ considerably differs between two piles of nails due to the different orientations, the rotationally-invariant distance $D_{\mathrm{S},\mc{R}}$ is more uniform. The bottom row displays pixelwise the optimal rotation angle that determines $D_{\mathrm{S},\mc{R}}$. Up to unavoidable local errors of these locally computed estimates, the distance $D_{\mathrm{S},\mc{R}}$ recovers the local orientation of the real texture in the input image (bottom right). 
} \label{fig:cov_table_dist}
\end{figure}

The six panels on the right of Figure \ref{fig:cov_texture_clustering} show columnwise the results of local label assignments ($k$-center + NN) and the assignments after label evolution performed by \textbf{(CFa)}, respectively, using either distance $D_{\mathrm{S}}$ or $D_{\mathrm{S},\mc{R}}$. Regarding the results depicted by the center column, greedy $k$-center clustering was performed using $D_{\mathrm{S},\mc{R}}$, while the nearest neighbor (NN) assignment and \textbf{(CFa)} were performed using $D_{\mathrm{S}}$, in order to highlight the difference between $D_{\mathrm{S}}$ and $D_{\mathrm{S},\mc{R}}$ based on the same initial prototypes.

The result shows that using $D_{\mathrm{S},\mc{R}}$ leads to an unsupervised labeling of all textures with a single label only. Thus, depending on the application, using $D_{\mathrm{S},\mc{R}}$ instead of the basic Stein divergence $D_{\mathrm{S}}$ can lead to more compact label dictionaries determined by the proposed unsupervised assignment flow. Figure \ref{fig:cov_texture_dist} underlines this finding from a different angle. The two panels on the left display \textit{pixelwise} the distances $D_{\mathrm{S}}$ and $D_{\mathrm{S},\mc{R}}$ to some fixed (arbitrary) reference descriptor. The two images show quantitatively that $D_{\mathrm{S}}$ is highly non-uniform, unlike $D_{\mathrm{S},\mc{R}}$. The panel on the right of Figure \ref{fig:cov_texture_dist} visualizes for each pixel the optimal angle minimizing \eqref{eq:stein-rotation-inv} over \eqref{eq:R-subgroup}, that has to be determined for the evaluation of $D_{\mathrm{S},\mc{R}}$. One can clearly see how the rotations of the textures of the input image of Figure \ref{fig:cov_texture_clustering} are recovered. This may be useful for some applications as well.

Figure \ref{fig:cov_table_clustering} depicts a real-world image. We used $|\mc{N}|=5 \times 5$ neighborhoods for spatial averaging and $\varepsilon = 10^{-7}$ for the constant of \eqref{eq:covariance-descriptor} to ensure strict positive definiteness of the covariance descriptors. Analogous to Figure \ref{fig:cov_texture_clustering}, we compared the nearest neighbor (NN) assignment and the result returned by \textbf{(CFa)} with respect to the Stein divergence $D_{\mathrm{S}}$ and its rotationally invariant version $D_{\mathrm{S},\mc{R}}$, respectively. 

We observe that the rotationally invariant feature representation together with the unsupervised assignment flow ($D_{\mathrm{S},\mc{R}}$ / \textbf{(CFa)}; panel bottom-right) leads to an unsupervised label representation of the input data that basically partitions the image into wooden texture independent of the orientation of the wooden boards (encoded with red), nails and similar line structures in the background (encoded with green), the hammers (light-blue) and oriented wooden texture (blue). 

Analogous to Figure \ref{fig:cov_texture_dist}, Figure \ref{fig:cov_table_dist} (first row) shows the pixelwise distances to a fixed label (located at the right pile of nails) for the distances $D_{\mathrm{S}}$ and $D_{\mathrm{S},\mc{R}}$, respectively. Comparing the distances to the two piles of nails illustrates once again and quantitatively the rotational invariance of $D_{\mathrm{S},\mc{R}}$. The bottom row of panels shows the corresponding optimal rotation angles corresponding to the evaluation of $D_{\mathrm{S},\mc{R}}$, as defined by \eqref{eq:stein-rotation-inv}. These angles recover the relative orientation of the textures which may be useful for some applications.

\section{Conclusion}
\label{sec:Conclusion}

We proposed the unsupervised assignment flow for performing jointly label evolution on feature manifolds and spatially regularized label assignment to given feature input data. The approach alleviates the requirement for supervised image labeling to have proper labels at hand, because an initial set of labels can evolve and adapt to better values while being assigned to given data.

The derivation of our approach highlights that it  encompasses related state-of-the-art approaches to unsupervised learning: soft-$k$-means clustering and EM-based estimation of mixture distributions with distributions of the exponential family as mixture components (class-conditional feature distributions). We generalized these approaches to manifold-valued data and defined the unsupervised assignment flow by coupling label evolution with the assignment flow adopted from \cite{Astrom:2017ac}. We suggested greedy $k$-center clustering for determining an initial label set that works with linear complexity in any metric space and with fixed approximation error bounded from above.

The separation between feature evolution and spatial regularization through assignments enables the flexible application of our approach to various scenarios, provided some key operations (divergence function evaluation, exponential map) are computational feasible for the particular feature manifold at hand. We demonstrated this property for three different scenarios and showed that coupling the evolution of labels and assignments has beneficial effects in either direction. The approach involved two parameters whose role is well understood. As a consequence, the only essential parameter is the neighborhood size used for spatial regularization.

Our unsupervised learning approach is consistent in that the very same approach that is used for supervised labeling is used for label learning, without need to resort to approximate inference due to the complexity of learning, as is the case, e.g., for learning with graphical models.

A key property of our approach is the sparsifying effect of spatial assignment regularization on unsupervised label learning. Our future work will study this property in connection with label learning from the assignment flow itself, in terms of patches of assignments at coarser spatial scales. Furthermore, all experiments in this paper were conducted using uniform weights $(w_{ik})_{k \in \mc{N}_{i}}$ for the spatial regularization of assignments (cf.~Eq.~\eqref{eq:def-geom-mean}). Learning these weights from data in order to represent the spatial context of typical feature occurrences as prior knowledge has been studied recently \cite{Huhnerbein:2019aa}. Working out a mathematically consistent way to extend this approach to unsupervised scenarios, as studied in the present paper, defines an exciting modeling problem.

\bibliographystyle{amsalpha}
\bibliography{main}
\clearpage

\end{document}